\documentclass{article}
\usepackage{arxiv_2024,times}

\usepackage{amsmath,amsfonts,bm}

\def\eqref#1{equation~\ref{#1}}

\def\1{\bm{1}}

\DeclareMathAlphabet{\mathsfit}{\encodingdefault}{\sfdefault}{m}{sl}
\SetMathAlphabet{\mathsfit}{bold}{\encodingdefault}{\sfdefault}{bx}{n}

\pdfoutput=1

\usepackage[utf8]{inputenc} 
\usepackage[T1]{fontenc}    
\usepackage{hyperref}       
\usepackage{url}            
\usepackage{booktabs}       
\usepackage{amsfonts}       
\usepackage{nicefrac}       
\usepackage{microtype}      
\usepackage{multirow}
\usepackage{amsmath}
\usepackage{tcolorbox}
\usepackage{cleveref}       
\usepackage{pifont}
\newcommand{\cmark}{\ding{51}}%
\newcommand{\xmark}{\ding{55}}%
\usepackage{adjustbox}
\usepackage{xspace}
\usepackage{enumitem}
\usepackage{wrapfig}
\usepackage{graphicx}

\newcommand\mypara[1]{\vspace{1.0mm}\noindent\textbf{#1}}
\newcommand{\ourmethod}{{\sc {EvalAlign}}\xspace}
\newcommand{\ourmethodbold}{{\textbf{\textsc {EvalAlign}}}\xspace}

\hypersetup{
    colorlinks,
    pagebackref=true,
    linkcolor={red!50!red},
    citecolor={citecolor!50!citecolor},
    urlcolor={blue!80!black}
}

\definecolor{citecolor}{RGB}{30,130,255}
\definecolor{OliveGreen}{rgb}{0,0.6,0}
\arxivfinalcopy

\title{EvalAlign: Supervised Fine-Tuning \\ Multimodal LLMs with Human-Aligned Data for Evaluating Text-to-Image Models}

\author{
  Zhiyu Tan$^1$ \quad
  Xiaomeng Yang$^2$ \quad
  Luozheng Qin$^2$ \quad
  Mengping Yang$^2$ \quad
  \textbf{Cheng Zhang}$^3$ \quad
  \textbf{Hao Li}$^1$ \\ [8pt]
  $^1$ Fudan University \quad 
  $^2$ Shanghai Academy of AI for Science \quad 
  $^3$ Carnegie Mellon University \\ [8pt]
  \url{https://sais-fuxi.github.io/projects/evalalign} \\
}
\newcommand{\smallcite}[1]{{\fontsize{7}{8}\selectfont\citep{#1}}}
\newcommand{\noSignal}{\textcolor{red}{\xmark}}
\newcommand{\yesSignal}{\textcolor{OliveGreen} {\bf \cmark}}

\begin{document}

\maketitle

\begin{abstract}
The recent advancements in text-to-image generative models have been remarkable. Yet, the field suffers from a lack of evaluation metrics that accurately reflect the performance of these models, particularly lacking fine-grained metrics that can guide the optimization of the models. In this paper, we propose \ourmethod, a metric characterized by its accuracy, stability, and fine granularity. Our approach leverages the capabilities of Multimodal Large Language Models (MLLMs) pre-trained on extensive data. We develop evaluation protocols that focus on two key dimensions: image faithfulness and text-image alignment. Each protocol comprises a set of detailed, fine-grained instructions linked to specific scoring options, enabling precise manual scoring of the generated images. We supervised fine-tune (SFT) the MLLM to align with human evaluative judgments, resulting in a robust evaluation model. Our evaluation across 24 text-to-image generation models demonstrate that \ourmethod not only provides superior metric stability but also aligns more closely with human preferences than existing metrics, confirming its effectiveness and utility in model assessment.
\end{abstract}
\section{Introduction}
\label{sec:intro}

Text-to-image models, such as DALL·E series~\citep{dalle2, dalle3}, Imagen~\citep{imagen}, and Stable Diffusion~\citep{sdxl}, have significantly impacted various domains such as entertainment, design, and education, by enabling high-quality image generation. 
These technologies not only advance the field of text-to-image generation but also bloom applications such as video generation~\citep{blattmann2023stable_video_diffusion, zhang2023i2vgen, tan2024vidgen}, image editing~\citep{sde,composer,controlnet} and human image generation~\citep{wang2024towards}.
Despite achieving incredible progress, the evaluation methods in this area are far from flawless and suffer heavily from data bias, as they are mainly trained on real images but are employed to evaluate synthesized images.

Since human-based evaluations are considerably costly in money and time, existing evaluation methods are primarily based on pretrained models, which are trained on real images.
However, the trained real images are generated by humans and high in image faithfulness and text-image alignment because of their generation essence.
Meanwhile, the evaluated images are synthesized by text-to-image models and encounter problems such as low image faithfulness or text-image alignment, constrained by the performance of generative models.

We dub the gap between the training data and the evaluated data as data bias, which may cause the evaluation models perform ill-suited on text-to-image evaluation.
Because of the data bias, existing text-to-image evaluation methods performs poorly in synthesized image evaluations.
Unfortunately, during our preliminary observation, nearly every synthesized images contain visual elements with low image faithfulness or text-image alignment, emphasize their significance on evaluation performance. 
Notably, there are also some works such as HPSv2~\citep{wu2023hps} and PickScore~\cite{kirstain2024pickscore}, where their evaluation models are trained synthesized images.
However, in their evaluation settings, the utilized synthesized images are treated as real images as they don't explicitly recognize the problem of synthesized images with low image faithfulness.

In view of these issues, we propose \ourmethod, a comprehensive, fine-grained and interpretable metric on text-to-image model assessing with low cost but high accuracy.
To build \ourmethod, we first curate a dataset composed of fine-grained human feedback scores on synthesized images, with consideration of the corresponding prompts.
The granularity of the feedback covers 11 skills categorized into two aspects: image faithfulness and text-image alignment.
After that, we Supervised finetune (SFT) a Multimodal Large Language Model (MLLM) on the annotated dataset, aligning it with human prior on detailed and accurate text-to-image evaluation.

Owing to extensive pre-training and large model capacity, MLLMs demonstrate excellent image-text understanding and generalization capabilities.
However, since the pre-training data does not include synthesized images with low image faithfulness or evaluation-related text instructions, using MLLMs directly for model evaluation may yield non-optimal results. 
Especially, we want to use MLLMs to support comprehensive and detailed evaluations, encompassing 11 skills and 2 aspects.
The definitions and nuances of these may not be fully understood by the MLLM.
Therefore, we employ SFT on a small amount of high-quality annotated data to align the MLLM with human judgement on evaluating synthesized images in criteria of 11 skills and 2 aspects.
Notably, since the main intelligence of \ourmethod stems from the annotated dataset and the utilized MLLM, we will make them accessible to the public. 

In summary, our main contributions can be summarized as follows:
\begin{itemize}[itemsep=1pt,topsep=0.5pt,leftmargin=12pt]

    \item We build a detailed human feedback dataset specifically designed to address the aforementioned challenges of text-to-image model evaluations. The annotated dataset is thoroughly cleaned, carefully balanced in topics, and systematically annotated by human. The dataset is composed by fine-grained human prior on evaluating synthesized images in criteria of 11 skills and 2 aspects.
    
    \item We propose \ourmethod, a text-to-image evaluation method which accurately aligns evaluation models with fine-grained human prior using the annotated dataset. \ourmethod exclusively supports an accurate, comprehensive, fine-grained and interpretable text-to-image evaluations. Besides \ourmethod is cost-effective in terms of annotation and training and computationally efficient.

    \item With \ourmethod, we conduct evaluations over 24 text-to-image models and compare \ourmethod with existing evaluation methods. Quantitative and qualitative experiments demonstrate that \ourmethod outperforms other methods in evaluating model performance.
 
\end{itemize}

\section{Related Work}
\label{sec:related_work}
\subsection{Benchmarks of text-to-image generation}
Despite the incredible progress achieved by text-to-image generation~\cite{zhang2023itigen, tan2024empirical}, evaluations and benchmarks in this area are far from flawless and contain critical limitations.
For example, the most commonly used metrics, IS~\citep{salimans2016is}, FID~\citep{heusel2017fid}, and CLIPScore~\citep{hessel2021clipscore} are broadly recognized as inaccurate for their inconsistency with human perception.
To address, HPS series~\citep{wu2023hps, wu2023hpsv2}, PickScore~\citep{kirstain2024pickscore}, and ImageReward~\citep{xu2024imagereward} introduced human preference prior on image assessing to the benchmark, thereby allowing better correlation with image quality.
However, with varying source and size of training data, these methods merely score the evaluated images in a coarse and general way, which cannot serve as an indication for model evolution.
Meanwhile, HEIM~\citep{lee2024heim} combined automatic and human evaluation and holistically evaluated text-to-image generation in 12 aspects, such as alignment, toxicity, and so on. 
As a consequence, HEIM relies heavily on human labour, limiting its application within budget-limited research groups severely.
\cite{otani2023toward} standardized the protocol and settings of human evaluation, ensuring its verifiable and reproducible.
Considering the issues of existing benchmarks, we propose \ourmethod to offer a cost-efficient, comprehensive and fine-grained text-to-image model evaluation.
Through our observations, we found that image faithfulness and text-image alignment are two key factors for comprehensive evaluation.
Image faithfulness requires the model to generate visual elements that are consistently faithful to the real-world.
For example, visual elements such as distorted body.
Meanwhile, text-image alignment measures how the generated images are aligned with their corresponding prompts.

There are also some works bear a resemble with us.
For instance, TIFA~\citep{hu2023tifa}, Gecko~\citep{wiles2024gecko} and LLMScore~\citep{lu2024llmscore} also formulate the evaluation as a set of visual question answering procedure and use LLMs as evaluation models.
However, while they all mainly focus on text-image alignment, our approach takes both text-image alignment and image faithfulness into consideration.
Moreover, the evaluation of LLMScore requires an object detection stage, which introduces significantly extra inference latency to the evaluation pipeline. 

As illustrated in Table~\ref{tab:benchmark_comparison}, existing text-to-image evaluation methods contains various limitations, making them incapable to serve as a fine-grained, comprehensive, and human-preference aligned automatic benchmark.
While our work fills in this gap economically, and can be employed to indicate evolution direction and support thorough analysis of text-to-image generation models.

\begin{table}[t]
    \centering
    \small
    \tabcolsep 3pt 
    \caption{\small \textbf{Comparison of different evaluation metrics and frameworks for text-to-image generation.} \ourmethod focuses on two key evaluation aspects, i.e., image faithfulness and text-image alignment, and supports human-aligned, fine-grained, and automatic evaluations. P: Prompt. I: Image. A: Annotation.}
    \resizebox{\textwidth}{!}{
    \begin{tabular}{l|r|ccc|ccc|cc}
    \toprule
    \multicolumn{1}{c}{\multirow{2.5}{*}{Method}} & \multicolumn{1}{|c}{\multirow{2.5}{*}{Venue}} & \multicolumn{3}{|c|}{Benchmark Feature} & \multicolumn{3}{c|}{Dataset Size} & \multicolumn{2}{c}{Evaluation Aspect} \\
    \cmidrule{3-10}
     & \multicolumn{1}{|c|}{} & Human-aligned & Fine-grained & Automatic & P & I & A & Faithfulness & Alignment \\ 
     \midrule
    Inception Score~\citep{salimans2016is} & NeurIPS 2016 & \noSignal  & \noSignal  & \yesSignal & -- & 1.3M & -- & \yesSignal & \noSignal \\
    FID~\citep{heusel2017fid} & NeurIPS 2017& \noSignal  & \noSignal  & \yesSignal & -- & 1.3M & -- & \yesSignal & \noSignal \\
    CLIP-score~\citep{hessel2021clipscore} & EMNLP 2021 & \noSignal  & \noSignal  & \yesSignal & 400M & 400M & -- & \noSignal & \yesSignal \\
    HPS~\citep{wu2023hps} & ICCV 2023& \yesSignal & \noSignal & \yesSignal & 25K & 98K & 25K & -- & -- \\
    TIFA~\citep{hu2023tifa} & ICCV 2023 & \yesSignal & \yesSignal & \yesSignal & 4K & -- & 25K & \noSignal &\yesSignal \\
    TVRHE~\citep{otani2023toward} & CVPR 2023& \yesSignal & \noSignal & \noSignal & -- & -- & -- & \yesSignal & \noSignal \\
    ImageReward~\citep{xu2024imagereward} & NeurIPS 2023 & \yesSignal & \noSignal & \yesSignal & 8.8K & 68K & 137K & -- & -- \\
    PickScore~\citep{kirstain2024pickscore} & NeurIPS 2023 & \yesSignal & \noSignal & \yesSignal & 35K & 1M & 500K & -- & -- \\
    HPS v2~\citep{wu2023hpsv2} &arXiv 2023& \yesSignal & \noSignal & \yesSignal & 107K & 430K & 645K & -- & -- \\
    HEIM~\citep{lee2024heim} & NeurIPS 2023& \yesSignal & \yesSignal & \noSignal & -- & -- & -- & \yesSignal & \yesSignal \\
    Gecko~\citep{wiles2024gecko} & arXiv 2024& \yesSignal & \yesSignal & \yesSignal & 2K & -- & 108K & \noSignal & \yesSignal \\
    LLMScore~\citep{lu2024llmscore} & arXiv 2024& \yesSignal & \yesSignal & \yesSignal & -- & -- & -- & \noSignal & \yesSignal \\ \midrule
    \ourmethod (ours) & \multicolumn{1}{c|}{--} & \yesSignal & \yesSignal & \yesSignal & 3K & 21K & 132K & \yesSignal & \yesSignal \\
    \bottomrule
    \end{tabular}
    }
    \label{tab:benchmark_comparison}
\end{table}

\subsection{Multimodal Large Language Models~(MLLMs)}
Pre-trained on massive text-only and image-text data, MLLMs have exhibited exceptional image-text joint understanding and generalization abilities, facilitating a large spectrum of downstream applications.
Among the works major in MLLMs, LLaVA~\citep{llava, llava-v1.5} and MiniGPT4~\citep{minigpt4, minigpt4-v2} observed that multimodal SFT is sufficient to align MLLMs with human preferences and enable them to accurately answer fine-grained questions about visual content.
Besides, Video-LLaMA~\citep{zhang2023video-llama} and VideoChat~\citep{li2023videochat} utilized  MLLMs for video understanding.
VILA~\citep{lin2023vila} quantitatively proved that involving text-only instruction-tuning data during SFT can further ameliorate model performance on text-only and multimodal downstream tasks.
LLaVA-NeXT~\citep{liu2024llavanext} extracted visual tokens for both the resized input image and the segmented sub-images to provide more detailed visual information for  MLLMs, achieving significant performance bonus on tasks with high-resolution input images.

However, due to the data bias, existing MLLMs cannot perfectly quantify for text-to-image evaluations.
Thus, we meticulously curate a SFT dataset to align MLLMs with detailed human feedback on synthesized images.

\section{\ourmethod Dataset Construction}
\label{sec:data_construction}

To train, validate and test the effectiveness of our evaluation models, we build \ourmethod dataset.
Specifically, \ourmethod dataset is a meticulously annotated collection featuring fine-grained annotations for images generated on text conditions. 
This dataset comprises 21k images, each accompanied by detailed instructions. 
The compilation process for the \ourmethod Dataset encompasses prompt collection, image generation, and precise instruction-based annotation.

\subsection{Prompts and Images Collection}
\label{subsec:data_collection}

\mypara{Prompt collection.}  
To assess the capabilities of our model in terms of image faithfulness and  text-image alignment, we collect, filter, and clean prompts from existing evaluation datasets and generated prompts based on LLM. 
These prompts encompass a diverse range from real-world user prompts, prompts generated through rule-based templates with LLM, to manually crafted prompts.
Specifically, the utilized prompts are soureced from HPS~\citep{wu2023hps}, HRS-Bench~\citep{bakr2023hrs-bench}, HPSv2~\citep{wu2023hpsv2}, TIFA~\citep{hu2023tifa}, DSG~\citep{cho2023dsg}, T2I-Comp~\citep{huang2023t2i-compbench}, Winoground~\citep{thrush2022winoground}, DALL-EVAL~\citep{cho2023dall-eval}, DiffusionDB~\citep{wang2023diffusiondb}, PartiPrompts~\citep{yu2022partiprompts}, DrawBench~\citep{imagen}, and JourneryDB~\citep{sun2024journeydb}.

\mypara{Prompt curation.}
To facilitate a clean and reasonable evaluation, each prompt to be annotated have to instruct text-to-image models to generate images that can reflect model performances on image faithfulness and text-image alignment.
However, considering some of the collected prompts fail to achieve the purpose, we need to filter and balance the collected prompts to ensure their quantity, quality and diversity.
For image faithfulness evaluation, we prioritize prompts related to human, animals, and other tangible objects, as prompts depicting sci-fi scenarios are less suitable for this type of assessment. 
Consequently, the prompt filter for image faithfulness initially selects prompts that describe human, animals, and other real objects. 
After deduplicating these prompts, we carefully select 1,500 distinct prompts with varying topic, background and style. The selected prompts encompass 10k subjects across 15 categories.
For text-image alignment evaluation, we refine our selection based on descriptions of style, color, quantity, and spatial relationships in the prompts. 
Specifically, only prompts contain relevant descriptions and exceed 15 words in length are considered, culminating in a final set of 1,500 prompts.

\begin{figure}[t]
\centering
\centerline{\includegraphics[width=1\linewidth]{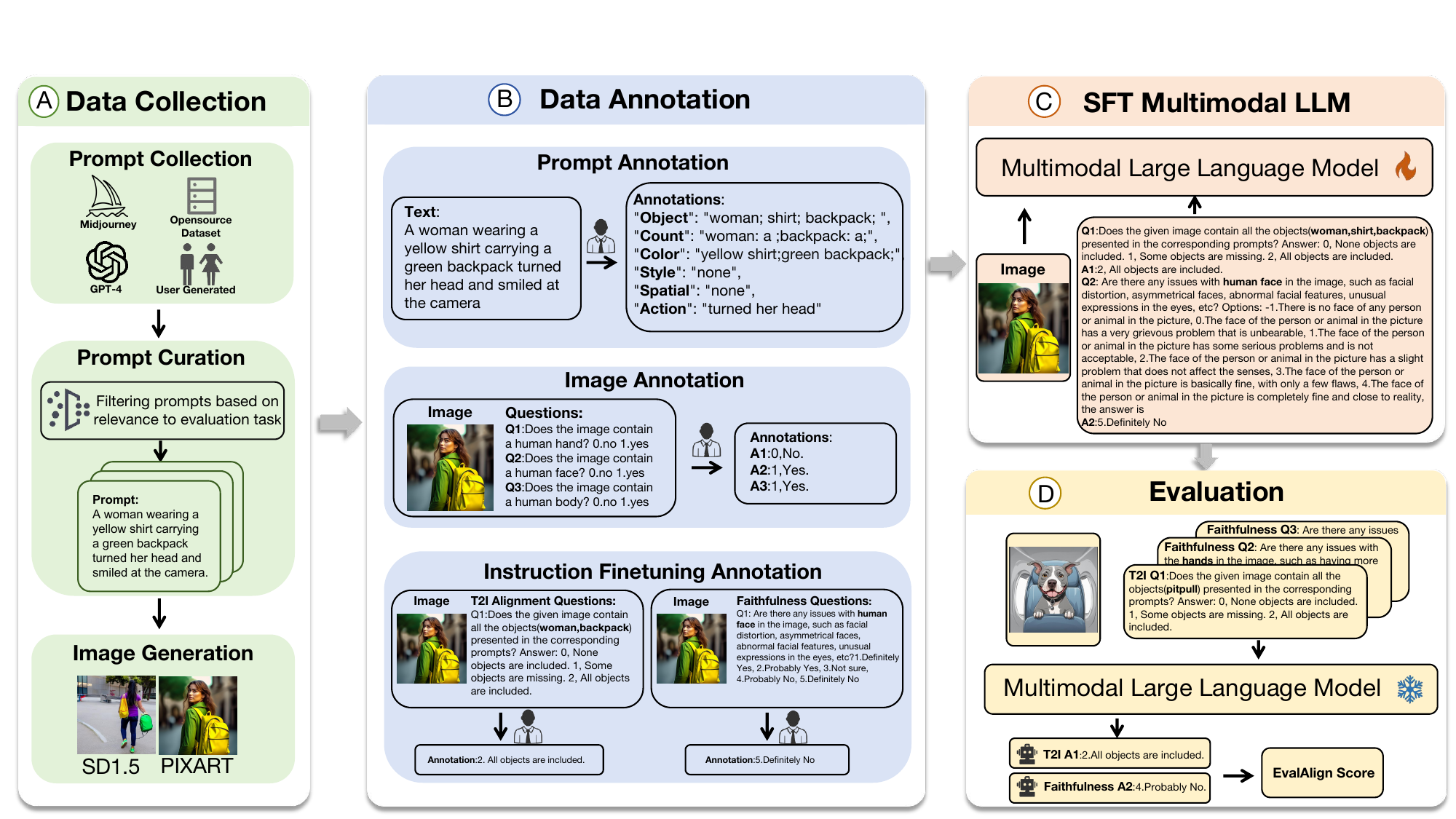}}
\vspace{-3mm}
\caption{
\small\textbf{Overview of \ourmethodbold.} We collect, filter and clean prompts from various sources to ensure their quantity, quality and diversity. We use 8 state-of-the-art text-to-image models to the generate images for evaluation. These synthesized images are then delegated to human annotators for thorough multi-turn annotation. Finally, the annotated data are used to finetune a MLLM to align it with fine-grained human preference, thereby adapting the model to perform text-to-image evaluation on image faithfulness and text-image alignment.
}
\label{fig:pipeline}
\vspace{-4mm}
\end{figure}

\mypara{Image generation.}
To train and evaluate the MLLM, we use a diverse set of images generated by various models using the aforementioned prompts, facilitating detailed human annotation. 
For each prompt, multiple images are generated across different models. 
The models used to generate these images vary in architectures and scales, enhancing the dataset diversity.
There are 24 models used to generate these images, varying in architecture as well as scale and thus enhancing the dataset diversity. 
For detailed information on the generation setting of each model, please refer to the appendix. 

The training and validation set comprises synthesized images from 8 out of the 24 models, whereas the test set spans all of them.
Particularly, the exclusive inclusion of the 16 models in the test set is crucial for validating the MLLM's ability to generalize beyond its training data.
Through our manual inspection, in this way, we attain ample synthesized images with a balanced diversity in the performance of image faithfulness and text-image alignment. 

\subsection{Data Annotation} 
\label{subsec:data_annotation}
\mypara{Prompt annotation.} 
For text prompts focused on text-image alignment, we begin by annotating the entities and their attributes within the text, as illustrated in Figure~\ref{fig:pipeline}.
Our annotators extract the entities mentioned in the prompts and label each entity with corresponding attributes, including quantity, color, spatial relationships, and actions.
During the annotation, we also ask the annotators to annotate the overall style of the image if described in the corresponding prompt and report prompts that contain toxic and NSFW content.
These high-quality and detailed annotations facilitate the subsequent SFT training and evaluation of the MLLM. 
The prompt annotation procedure ensures that the MLLM can accurately align and respond to the nuanced details specified in the prompts, enhancing both the training process and the model's performance in generating images that faithfully reflect the described attributes and style.

\mypara{Image annotation.} 
The images generated by text-to-image models often present challenges such as occluded human body parts, which can impede the effectiveness of SFT training and evaluation of the MLLM. 
To address these challenges and enhance the model's training and evaluative capabilities, specific annotations are applied to all images depicting human and animals. 
These annotations include: presence of human or animal faces; visibility of hands; visibility of limbs.
By implementing these annotations, we ensure that the MLLM  can more effectively learn from and assess the completeness and faithfulness of the generated images. 
This structured approach to annotation not only aids in identifying common generation errors but also optimizes the model's ability to generate more accurate and realistic images, thereby improving both training outcomes and the model's overall performance in generating coherent and contextually appropriate visual content.

\mypara{Instruction-finetuning data annotation.} 
To align the MLLM with human preference prior on detailed synthesized image assessing, we can train the model on a minimal amount of fine-grained human feedback data through SFT training.
As a consequence, we devise two sets of questions, each is concentrated on a specific fine-grained skill of image faithfulness and image-text alignment.
Human annotators are required to answer these questions to acquire the fine-grained human preference data.
To aid them to understand the meaning and principle of each question,  thereby ensuring high annotation quality, we employ a thorough and comprehensive procedure of annotation preparation.
First, we write a detailed annotation guideline and conduct a training for the annotators to explain the annotation guideline and answer their questions about the annotation.
Then, we conduct a multi-turn trial annotation on another 50 synthesized images.
After each trial, we calculate the Cohen's kappa coefficient and interpret annotation guidelines for our annotators.
In total, we conduct nine turns of trial annotation, and in the last turn of the trial, the Cohen's kappa coefficient of our annotators reaches $0.681$, indicating high inter-annotator reliability and high annotation quality.

After completing the aforementioned preparations, we delegate the images filtered during image annotation to 10 annotators and ask them to complete the annotation just as how they did in the trial annotation.
Furthermore, during the whole annotation procedure, four experts in text-to-image generation conduct random sampling quality inspection on the present annotated results, causing a second and a third re-annotation on 423 and 112 inspection-failed samples.
Overall, owing to the valuable work of our human annotators and our fastidious annotation procedure, we get quality-sufficient instruction-tuning data required for the SFT training of the MLLM.
More details of the annotation procedure will be introduced in supplementary files.

\subsection{Dataset Statistics}
\label{subsec:statistics}
\begin{figure}[t]
  \centering
  \begin{minipage}{0.48\textwidth}
    \includegraphics[width=1\linewidth]{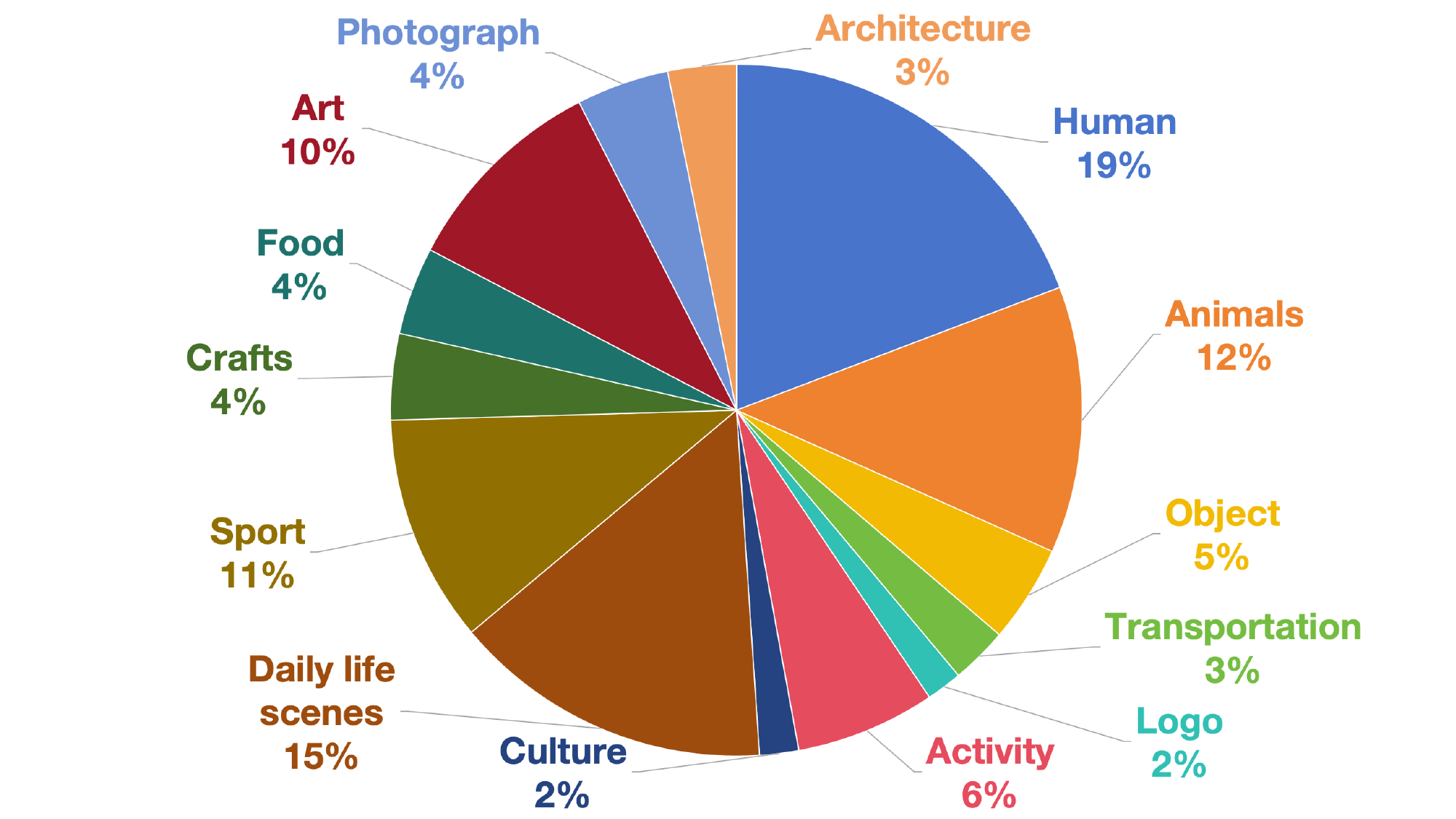}
    \caption{\small
    \textbf{Statistics of prompts on evaluating text-to-image alignment.} Prompts in our text-to-image alignment benchmark covers a broad range of concepts commonly used in text-to-image generation.}
    \label{fig:alignment_prompts}
  \end{minipage}\hfill
  \begin{minipage}{0.48\textwidth}
    \includegraphics[width=1\linewidth]{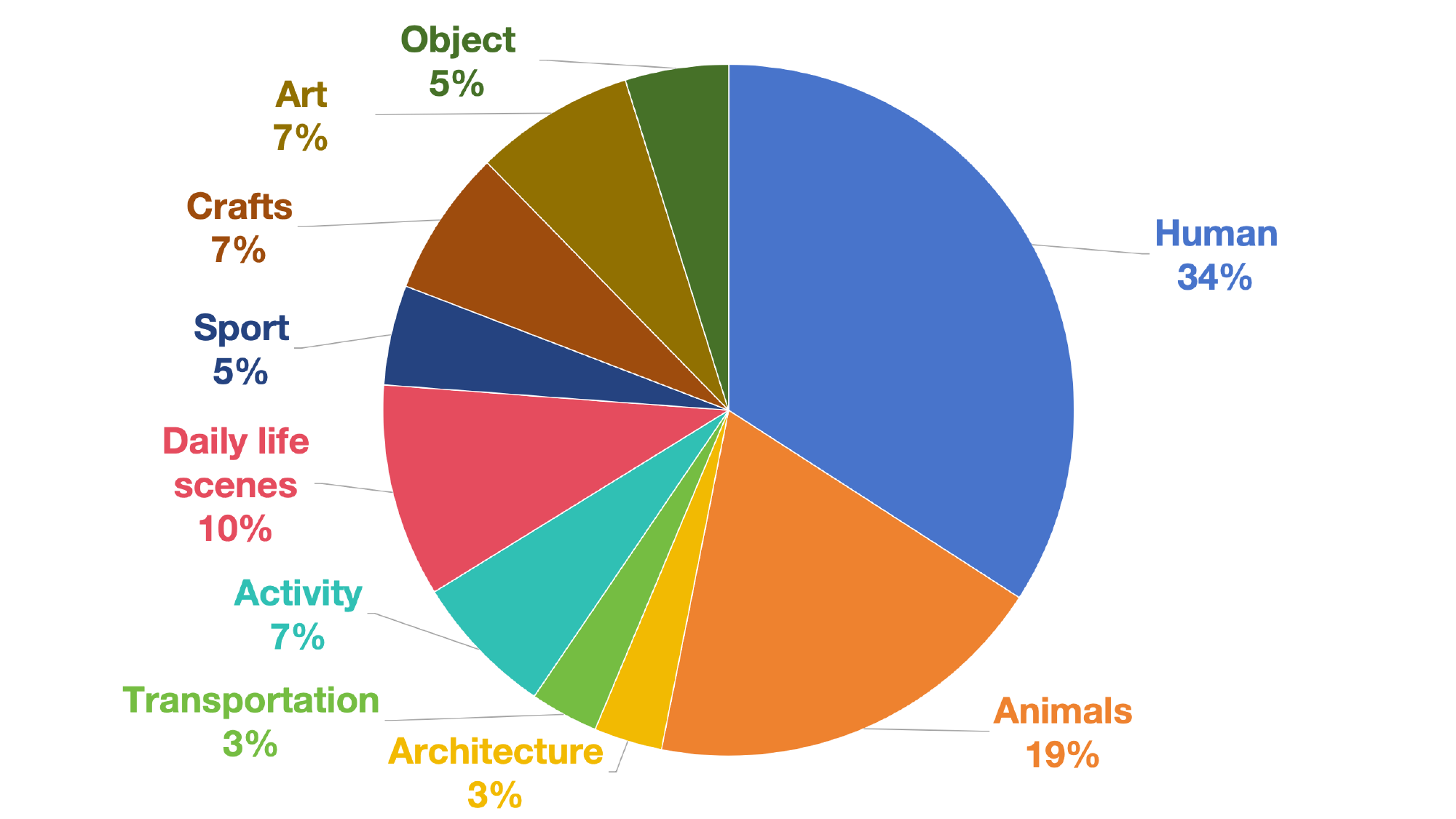}
    \caption{\small
    \textbf{Statistics of prompts on evaluating image faithfulness.} Prompts in our image faithfulness benchmark covers a broad range of objects and categories that related to image faithfulnes.
    }
    \label{fig:faithfulness_prompts}
  \end{minipage}
\end{figure}

To summarize, we generate 24k images from 3k prompts based on 8 text-to-image models, which includes DeepFloyd IF~\citep{DeepFloydIF}, SD15~\citep{Rombach_2022_sd15_sd21}, LCM~\citep{luo2023lcm}, SD21~\citep{Rombach_2022_sd15_sd21}, SDXL~\citep{sdxl}, Wuerstchen~\citep{pernias2023wuerstchen}, Pixart~\citep{chen2023pixart}, and SDXL-Turbo~\citep{sdxl-turbo}. 
After data filtering, 4.5k images are selected as annotation data for task of text-image alignment. 
Subsequently, these images are carefully annotated to generate 13.5k text-image pairs, where 11.4k are used to the training dataset and 2.1k to the validation dataset. 
For the image faithfulness task, we select 12k images for annotation, yielding 36k text-image pairs, with 30k are used to the training dataset and 6.2k to the validation dataset.
Additionally, we employed 24 text-to-image models to generate 2.4k images from 100 prompts. 
After annotation, these images are used as testing dataset. 
Figure~\ref{fig:alignment_prompts} and Figure~\ref{fig:faithfulness_prompts} show the distribution of objects in different categories within our prompts, demonstrating the diversity and balance of our prompts.

\section{Training and Evaluation Methods}

\subsection{Supervised Finetuning the MLLM}
As we mentioned above, we use MLLMs as the evaluation models and let it to answer a set of carefully-designed instructions, thereby achieving quantitative measurement of fine-grained text-to-image generation skills.
Due to data bias, zero-shot MLLMs perform poorly when it comes to evaluation on generated images, particularly in term of image faithfulness.
To solve this problem, we apply SFT training on the detailed human annotation to align the MLLM with human preference prior.
Formally, the SFT training sample can be denoted as a triplet: question (or the instruction), multimodal input and answer.
During SFT training, the optimization objective is the autoregressive loss function utilized to train LLMs, but calculated only on the answer, the loss function can be formulated as follows:
\begin{align}
    L(\theta) = \sum^N_{i=1} \log p(A_i|Q,M,A_{<i};\theta),
\end{align}
where $N$ is the length of the ground truth answer, $Q$ is a fine-grained question of the generated image and its available answer, $M$ is the image and textual description, while $A$ is the human annotated answer selected from the given options.
Notably, we expand each option to make it more detailed and descriptive, thereby benefiting SFT performance by allowing the MLLM to better understand the meaning of each option.

\subsection{Evaluation and Metrics}
To evaluate synthesized images with consideration of its synthetic nature, \ourmethod is designed to evaluate image faithfulness and text-image alignment in a fine-grained way.
Notably, image faithfulness and text-image alignment are two common errors occurred in synthesized images, whereas real images inherently exhibit high levels of both image faithfulness and text-image alignment.

\mypara{Image Faithfulness} measures whether synthesized images are faithful to real-world commonsense. With higher image faithfulness, the visual elements of generated images more closely resemble their real-world counterparts. Unfortunately, text-to-image models often generate images with low faithfulness, such as distorted body structures and human hands. This is also a critical reason why we set image faithfulness as one of the benchmarking aspects in \ourmethod.
Additionally, evaluating image faithfulness requires considering the input prompts, as prompts may describe unreal or impossible scenarios that inherently affect the faithfulness of the generated images. For example, when prompts like "a dog walking like a human" or "a man on Mars without a spacesuit" are provided, the generated images may naturally deviate from real-world image faithfulness. Under such circumstances, the synthesized images cannot be regarded as low in image faithfulness since the generative models are merely following prompts that contain super-reality scenarios.

\mypara{Text-Image Alignment} evaluates whether generated images are aligned with their conditioned prompts. In the inference settings of text-to-image models, the image generation process is conditioned on textual prompts, requiring alignment between the text prompts and the synthesized images. However, through our observations, text-to-image models cannot consistently follow input prompts, often yielding images with visual elements misaligned with the input prompts. For example, models may generate images featuring an orange cat when conditioned on the text prompt "a blue cat."

During inference, the multimodal large language model (MLLM) is required to generate an appropriate response given a specific question $Q$ and multimodal input $M$ in an autoregressive manner:
\begin{align}
    R_i = f(Q, M, R_{<i};\theta),
\end{align}
where $R_i$ is the $i$-th generated token, $R_{<i}$ represents the sequence of tokens generated before step $i$, and $\theta$ denotes the parameters of the fine-tuned MLLM.
This autoregressive generation process is considered complete once the model generates an end-of-sequence (EOS) token or the generated response exceeds a preset maximum generation length.
After generation, we employ rule-based filtering and regular expressions to extract the option chosen by the MLLM. Each option is assigned a unique predefined score to quantitatively measure a fine-grained skill specified by the question $Q$:
\begin{align}
    \text{Score}(Q) = g(R) = g(f(Q,M;\theta)),
\end{align}
where $g(\cdot)$ represents the procedure of option extraction and score mapping.

We devise two holistic and detailed question sets, $S_{f}$ and $S_{a}$, that encompass every aspect of image faithfulness and text-image alignment, respectively.
Consequently, our metric, \textbf{EvalAlign}, can be defined by averaging the scores of the questions in the two sets:
\begin{align}
\text{EvalAlign}_{\text{f}} &= \frac{1}{|S_{f}|} \sum_{Q_i \in S_{f}} \text{Score}(Q_i), \\
\text{EvalAlign}_{\text{a}} &= \frac{1}{|S_{a}|} \sum_{Q_j \in S_{a}} \text{Score}(Q_j),
\end{align}
where $\text{EvalAlign}_{\text{f}}$ and $\text{EvalAlign}_{\text{a}}$ indicate the image faithfulness score and the text-image alignment score evaluated by our method, respectively.

\subsection{Implementation Details}
\label{subsec:implementation_details} 
For details about the SFT training, we apply LoRA~\citep{hu2021lora} finetuning on LLaVA-NeXT~\citep{liu2024llavanext} models to align them with the \ourmethod dataset.
Additionally, we merely adapt LoRA finetuning on the Q and K weights of the attention module, as extending the finetuning to the ViT~\citep{dosovitskiy2020vit} and projection modules will lead to overfitting.
The entire training process is conducted on $32$ NVIDIA A100 GPUs for $10$ hours, with a learning rate of 5$\times$10$^{-5}$.
As for the ablation study, we evaluate the finetuned LLaVA-NeXT 13B model on the validation dataset. 
In the final experiment, we apply SFT to the LLaVA-NeXT 34B model on the testing dataset to testify its generalization ability.

\section{Experimental Results}
\label{sec:experiments}
\subsection{Main Results}

\begin{table}[ht]
    \centering
    \small
    \fontsize{8}{8}\selectfont
    \tabcolsep 0.1pt 
    \caption{\small \textbf{Results on image faithfulness.} We evaluate the image faithfulness of images generated by 24 text-to-image models to compare five evaluation metrics against human scoring results. The experiments show that our metric's scores align more closely with human evaluations than those of other metrics.}
    \vspace{2mm}
    \label{tab:human alignment faithfulness}
    \begin{tabular}{lcccccc}
            \toprule
             Model & Human~~~~ & \ourmethod~~ & HPS v2 & CLIP-score & ImageReward  & PickScore \\
                \midrule
                PixArt XL2 1024 MS~\smallcite{chen2023pixart}                       & \colorbox{orange!100}{2.2848}$^{1~~}$  & \colorbox{orange!100}{1.6415}$^{1~~}$ & \colorbox{orange!100}{31.6226}$^{1~~}$ & \colorbox{orange!100}{0.8580}$^{1~~}$ & \colorbox{orange!100}{0.9696 }$^{1~~}$ & \colorbox{orange!100}{22.1335}$^{1~~}$ \\
                Dreamlike Photoreal v2.0~\smallcite{dreamlike-photoreal-20}         & \colorbox{orange!96}{2.0070}$^{2~~}$  & \colorbox{orange!88}{1.4522}$^{4~~}$  & \colorbox{orange!80}{29.2322}$^{6~~}$ & \colorbox{orange!56}{0.8286}$^{12}$ & \colorbox{orange!52}{0.1886 }$^{13}$ & \colorbox{orange!72}{21.2271}$^{8~~}$ \\
                SDXL Refiner v1.0~\smallcite{sdxlrefiner}                           & \colorbox{orange!92}{1.9229}$^{3~~}$  & \colorbox{orange!96}{1.6072}$^{2~~}$  & \colorbox{orange!92}{29.8197}$^{3~~}$ & \colorbox{orange!96}{0.8566}$^{2~~}$ & \colorbox{orange!96}{0.7245 }$^{2~~}$ & \colorbox{orange!96}{22.0492}$^{2~~}$ \\
                SDXL v1.0~\smallcite{sdxl}                                          & \colorbox{orange!88}{1.8136}$^{4~~}$  & \colorbox{orange!92}{1.4675}$^{3~~}$ & \colorbox{orange!76}{29.0620}$^{7~~}$ & \colorbox{orange!88}{0.8467}$^{4~~}$ & \colorbox{orange!92}{0.7043 }$^{3~~}$ & \colorbox{orange!92}{21.8106}$^{3~~}$ \\
                Wuerstchen~\smallcite{pernias2023wuerstchen}                        & \colorbox{orange!84}{1.7837}$^{5~~}$  & \colorbox{orange!84}{1.4279}$^{5~~}$  & \colorbox{orange!96}{30.6622}$^{2~~}$ & \colorbox{orange!48}{0.8199}$^{14}$ & \colorbox{orange!60}{0.3212 }$^{11}$  & \colorbox{orange!80}{21.3720}$^{6~~}$ \\
                LCM SDXL~\smallcite{luo2023lcm}                                     & \colorbox{orange!80}{1.6910}$^{6~~}$  & \colorbox{orange!76}{1.3391}$^{7~~}$  & \colorbox{orange!84}{29.3588}$^{5~~}$ & \colorbox{orange!64}{0.8335}$^{10}$ & \colorbox{orange!80}{0.5304 }$^{6~~}$ & \colorbox{orange!88}{21.6532}$^{4~~}$ \\
                Openjourney~\smallcite{openjourne}                                  & \colorbox{orange!76}{1.6667}$^{7~~}$   & \colorbox{orange!64}{1.1750}$^{10}$  & \colorbox{orange!52}{26.3475}$^{13}$ & \colorbox{orange!44}{0.8196}$^{15}$ & \colorbox{orange!40}{0.1478 }$^{16}$  & \colorbox{orange!64}{20.8637}$^{10}$ \\
                Safe SD MAX~\smallcite{SafeLatentDiffusion}                         & \colorbox{orange!72}{1.6491}$^{8~~}$  & \colorbox{orange!72}{1.2175}$^{8~~}$ & \colorbox{orange!36}{25.7396}$^{17}$ & \colorbox{orange!8}{0.7555}$^{24}$ & \colorbox{orange!16}{-0.0507}$^{22}$ & \colorbox{orange!20}{20.4594}$^{21}$ \\
                LCM LORA SDXL~\smallcite{luo2023lcm}                                & \colorbox{orange!68}{1.6387}$^{9~~}$  & \colorbox{orange!80}{1.3833}$^{6~~}$ & \colorbox{orange!64}{27.3299}$^{10}$ & \colorbox{orange!72}{0.8364}$^{8~~}$ & \colorbox{orange!76}{0.4959 }$^{7~~}$ & \colorbox{orange!84}{21.4824}$^{5~~}$ \\
                Safe SD STRONG~\smallcite{SafeLatentDiffusion}                     & \colorbox{orange!64}{1.6308}$^{10}$  & \colorbox{orange!60}{1.1466}$^{11}$  & \colorbox{orange!32}{25.5764}$^{18}$ & \colorbox{orange!32}{0.8165}$^{18}$ & \colorbox{orange!12}{-0.1022}$^{23}$ & \colorbox{orange!32}{20.6211}$^{18}$ \\
                Safe SD MEDIUM~\smallcite{SafeLatentDiffusion}                     & \colorbox{orange!60}{1.6275}$^{11}$  & \colorbox{orange!44}{1.1298}$^{15}$   & \colorbox{orange!48}{26.2798}$^{14}$  & \colorbox{orange!24}{0.8101}$^{20}$ & \colorbox{orange!56}{0.2042 }$^{12}$& \colorbox{orange!56}{20.7880}$^{12}$ \\
                Safe SD WEAK~\smallcite{SafeLatentDiffusion}                        & \colorbox{orange!56}{1.6078}$^{12}$  & \colorbox{orange!36}{1.1188}$^{17}$ & \colorbox{orange!44}{26.1180}$^{15}$ & \colorbox{orange!12}{0.7809}$^{23}$ & \colorbox{orange!8}{-0.1264}$^{24}$ & \colorbox{orange!8}{20.3873}$^{24}$ \\
                SD v2.1~\smallcite{Rombach_2022_sd15_sd21}                          & \colorbox{orange!52}{1.5524}$^{13}$  & \colorbox{orange!32}{1.1094}$^{18}$   & \colorbox{orange!56}{26.5823}$^{12}$ & \colorbox{orange!76}{0.8377}$^{7~~}$ & \colorbox{orange!68}{0.4116 }$^{9~~}$ & \colorbox{orange!68}{21.0502}$^{9~~}$ \\
                SD v2.0~\smallcite{Rombach_2022_sd15_sd21}                         & \colorbox{orange!48}{1.5277}$^{14}$ & \colorbox{orange!48}{1.1300}$^{14}$ & \colorbox{orange!20}{25.3481}$^{21}$ & \colorbox{orange!36}{0.8170}$^{17}$ & \colorbox{orange!32}{0.0872 }$^{18}$ & \colorbox{orange!52}{20.7529}$^{13}$ \\
                Openjourney v2~\smallcite{openjourneV2}                             & \colorbox{orange!44}{1.5000}$^{15}$  & \colorbox{orange!24}{0.9956}$^{20}$  & \colorbox{orange!12}{24.6984}$^{23}$ & \colorbox{orange!16}{0.7958}$^{22}$ & \colorbox{orange!20}{-0.0415}$^{21}$ & \colorbox{orange!16}{20.4088}$^{22}$ \\
                Redshift diffusion~\smallcite{redshiftdiffusion}                    & \colorbox{orange!40}{1.4733}$^{16}$ &  \colorbox{orange!56}{1.1382}$^{12}$  & \colorbox{orange!16}{25.1572}$^{22}$ & \colorbox{orange!20}{0.8101}$^{21}$ & \colorbox{orange!24}{0.0218 }$^{20}$ & \colorbox{orange!28}{20.6155}$^{19}$ \\
                Dreamlike Diffusion v1.0~\smallcite{dreamlike-diffusion-10}         & \colorbox{orange!36}{1.4652}$^{17}$  & \colorbox{orange!68}{1.2052}$^{9~~}$  &\colorbox{orange!88}{29.6506}$^{4~~}$ & \colorbox{orange!92}{0.8543}$^{3~~}$ & \colorbox{orange!88}{0.6508 }$^{4~~}$ & \colorbox{orange!76}{21.2664}$^{7~~}$ \\
                SD v1.5~\smallcite{Rombach_2022_sd15_sd21}                          & \colorbox{orange!32}{1.4417}$^{18}$  & \colorbox{orange!52}{1.1362}$^{13}$  & \colorbox{orange!28}{25.4972}$^{19}$ & \colorbox{orange!52}{0.8214}$^{13}$  & \colorbox{orange!48}{0.1686 }$^{14}$ & \colorbox{orange!40}{20.7143}$^{16}$ \\
                IF-I-XL v1.0~\smallcite{DeepFloydIF}                                & \colorbox{orange!28}{1.3808}$^{19}$  & \colorbox{orange!16}{0.9221}$^{22}$   & \colorbox{orange!68}{27.4512}$^{9~~}$ & \colorbox{orange!84}{0.8449}$^{5~~}$ & \colorbox{orange!84}{0.6087 }$^{5~~}$ & \colorbox{orange!48}{20.7474}$^{14}$ \\
                SD v1.4~\smallcite{Rombach_2022_sd15_sd21}                          & \colorbox{orange!24}{1.3592}$^{20}$   & \colorbox{orange!20}{0.9511}$^{21}$  & \colorbox{orange!24}{25.3697}$^{20}$  & \colorbox{orange!40}{0.8190}$^{16}$  & \colorbox{orange!36}{0.1050 }$^{17}$ & \colorbox{orange!36}{20.6535}$^{17}$ \\
                Vintedois Diffusion v0.1~\smallcite{vintedoisdiffusion}             & \colorbox{orange!20}{1.3562}$^{21}$  & \colorbox{orange!28}{1.0797}$^{19}$  & \colorbox{orange!60}{26.5901}$^{11}$ & \colorbox{orange!68}{0.8341}$^{9~~}$ & \colorbox{orange!64}{0.3562 }$^{10}$ & \colorbox{orange!60}{20.8358}$^{11}$ \\
                IF-I-L v1.0~\smallcite{DeepFloydIF}                                 & \colorbox{orange!16}{1.2635}$^{22}$  & \colorbox{orange!12}{0.8814}$^{23}$   & \colorbox{orange!72}{27.4836}$^{8~~}$ & \colorbox{orange!80}{0.8384}$^{6~~}$  & \colorbox{orange!72}{0.4463 }$^{8~~}$ & \colorbox{orange!44}{20.7170}$^{15}$  \\
                MultiFusion~\smallcite{MultiFusion}                                & \colorbox{orange!12}{1.2372}$^{23}$  & \colorbox{orange!40}{1.1298}$^{16}$ & \colorbox{orange!8}{23.8133}$^{24}$ & \colorbox{orange!28}{0.8151}$^{19}$ & \colorbox{orange!28}{0.0695 }$^{19}$ & \colorbox{orange!24}{20.4780}$^{20}$ \\
                IF-I-M v1.0~\smallcite{DeepFloydIF}                                & \colorbox{orange!8}{1.0135}$^{24}$  & \colorbox{orange!8}{0.7928}$^{24}$  & \colorbox{orange!40}{25.9522}$^{16}$  & \colorbox{orange!60}{0.8329}$^{11}$ & \colorbox{orange!44}{0.1637 }$^{15}$ & \colorbox{orange!12}{20.4035}$^{23}$ \\
                \bottomrule
          \end{tabular}
          \vspace{-5mm}
\end{table}

\begin{table}[ht]
    \centering
    \footnotesize
    \fontsize{8}{8}\selectfont
    \tabcolsep 0.1pt 
    \caption{\small \textbf{Results on text-to-image alignment. } We evaluated the text-image alignment of images generated by 24 text-to-image models to compare how five evaluation metrics align with human scoring results. The experiments reveal that, in terms of text-image alignment metrics, our metric scores are highly consistent with human scores, demonstrating a much closer alignment than other evaluation metrics. }
    \vspace{2mm}
    \label{tab:huamn alignment}
    \begin{tabular}{lcccccc}
                            \toprule
                Model & Human~~~~ & \ourmethod~~  & HPS v2  & CLIP-score & ImageReward  & PickScore  \\
                \midrule
                IF-I-XL v1.0~\smallcite{DeepFloydIF}                 & \colorbox{orange!100}{5.4500}$^{1~~}$  & \colorbox{orange!100}{5.5300}$^{1~~}$ & \colorbox{orange!64}{32.5477}$^{10}$ & \colorbox{orange!96}{0.8579}$^{2~~}$ & \colorbox{orange!92}{0.4391 }$^{3~~}$ & \colorbox{orange!64}{21.1998}$^{10}$ \\
                IF-I-L v1.0~\smallcite{DeepFloydIF}                  & \colorbox{orange!96}{5.2300}$^{2~~}$  &\colorbox{orange!96}{5.4500}$^{2~~}$   & \colorbox{orange!68}{32.7140}$^{9~~}$ & \colorbox{orange!88}{0.8538}$^{4~~}$ & \colorbox{orange!80}{0.3820 }$^{6~~}$ & \colorbox{orange!56}{21.1284}$^{12}$ \\
                SDXL Refiner v1.0~\smallcite{sdxlrefiner}            & \colorbox{orange!92}{5.2100}$^{3~~}$  & \colorbox{orange!92}{5.4000}$^{3~~}$ & \colorbox{orange!92}{35.6465}$^{3~~}$ & \colorbox{orange!84}{0.8528}$^{5~~}$ & \colorbox{orange!96}{0.4738 }$^{2~~}$ & \colorbox{orange!96}{22.3532}$^{2~~}$ \\
                LCM SDXL~\smallcite{luo2023lcm}                      & \colorbox{orange!88}{5.1800}$^{4~~}$  & \colorbox{orange!84}{5.3300}$^{5~~}$   & \colorbox{orange!80}{33.8011}$^{6~~}$ & \colorbox{orange!80}{0.8512}$^{6~~}$ & \colorbox{orange!84}{0.3833 }$^{5~~}$ & \colorbox{orange!88}{21.9620}$^{4~~}$ \\
                PixArt XL2 1024 MS~\smallcite{chen2023pixart}        & \colorbox{orange!84}{5.1100}$^{5~~}$  & \colorbox{orange!80}{5.3100}$^{6~~}$ & \colorbox{orange!100}{37.0493}$^{1~~}$ & \colorbox{orange!100}{0.8634}$^{1~~}$ & \colorbox{orange!100}{0.6542 }$^{1~~}$ & \colorbox{orange!100}{22.3926}$^{1~~}$ \\
                IF-I-M v1.0~\smallcite{DeepFloydIF}                  & \colorbox{orange!80}{5.0800}$^{6~~}$  & \colorbox{orange!72}{5.2200}$^{8~~}$ & \colorbox{orange!50}{31.0951}$^{14}$ & \colorbox{orange!72}{0.8434}$^{8~~}$ & \colorbox{orange!64}{0.0499 }$^{10}$  & \colorbox{orange!26}{20.8270}$^{20}$ \\
                LCM LORA SDXL~\smallcite{luo2023lcm}                 & \colorbox{orange!76}{5.0600}$^{7~~}$  & \colorbox{orange!76}{5.2700}$^{7~~}$ & \colorbox{orange!72}{32.7752}$^{8~~}$ & \colorbox{orange!64}{0.8349}$^{10}$ & \colorbox{orange!68}{0.1618 }$^{9~~}$ & \colorbox{orange!80}{21.7627}$^{6~~}$ \\
                SDXL v1.0~\smallcite{sdxl}                           & \colorbox{orange!72}{5.0300}$^{8~~}$  & \colorbox{orange!88}{5.3500}$^{4~~}$ & \colorbox{orange!88}{35.1593}$^{4~~}$ & \colorbox{orange!92}{0.8540}$^{3~~}$ & \colorbox{orange!88}{0.4322 }$^{4~~}$ & \colorbox{orange!92}{22.1291}$^{3~~}$ \\
                Wuerstchen~\smallcite{pernias2023wuerstchen}         & \colorbox{orange!68}{4.8700}$^{9~~}$  & \colorbox{orange!68}{5.1700}$^{9~~}$  & \colorbox{orange!96}{36.4632}$^{2~~}$ & \colorbox{orange!68}{0.8381}$^{9~~}$ & \colorbox{orange!76}{0.2513 }$^{7~~}$ & \colorbox{orange!84}{21.7779}$^{5~~}$ \\
                Openjourney~\smallcite{openjourne}                   & \colorbox{orange!64}{4.8300}$^{10}$  & \colorbox{orange!46}{4.9200}$^{15}$  & \colorbox{orange!56}{31.1495}$^{12}$ & \colorbox{orange!42}{0.8173}$^{16}$ & \colorbox{orange!50}{-0.0867}$^{14}$ & \colorbox{orange!54}{21.1163}$^{13}$ \\
                SD v2.1~\smallcite{Rombach_2022_sd15_sd21}           &  \colorbox{orange!60}{4.8000}$^{11}$    & \colorbox{orange!60}{5.0700}$^{11}$ & \colorbox{orange!54}{31.1017}$^{13}$ & \colorbox{orange!50}{0.8278}$^{14}$ & \colorbox{orange!56}{-0.0453}$^{12}$ & \colorbox{orange!68}{21.2093}$^{9~~}$ \\
                MultiFusion~\smallcite{MultiFusion}                  & \colorbox{orange!56}{4.6800}$^{12}$  & \colorbox{orange!34}{4.8000}$^{18}$  & \colorbox{orange!10}{28.7957}$^{24}$ & \colorbox{orange!46}{0.8264}$^{15}$ & \colorbox{orange!46}{-0.1337}$^{15}$ & \colorbox{orange!38}{20.9625}$^{17}$ \\
                Dreamlike Diffusion v1.0~\smallcite{dreamlike-diffusion-10}   & \colorbox{orange!54}{4.6600}$^{13}$  & \colorbox{orange!64}{5.1500}$^{10}$   & \colorbox{orange!84}{34.8196}$^{5~~}$ & \colorbox{orange!76}{0.8493}$^{7~~}$ & \colorbox{orange!72}{0.2295 }$^{8~~}$ & \colorbox{orange!76}{21.5550}$^{7~~}$ \\
                SD v2.0~\smallcite{Rombach_2022_sd15_sd21}                    & \colorbox{orange!50}{4.6400}$^{14}$  & \colorbox{orange!56}{5.0100}$^{12}$  & \colorbox{orange!38}{30.6153}$^{17}$ & \colorbox{orange!54}{0.8298}$^{13}$ & \colorbox{orange!42}{-0.1424}$^{16}$ &  \colorbox{orange!60}{21.1905}$^{11}$ \\
                Vintedois Diffusion v0.1~\smallcite{vintedoisdiffusion}       & \colorbox{orange!46}{4.6200}$^{15}$  & \colorbox{orange!50}{4.9500}$^{14}$  &  \colorbox{orange!60}{31.9503}$^{11}$  & \colorbox{orange!56}{0.8319}$^{12}$ &  \colorbox{orange!60}{-0.0222}$^{11}$ & \colorbox{orange!50}{21.1141}$^{14}$\\
                Safe SD STRONG~\smallcite{SafeLatentDiffusion}                & \colorbox{orange!42}{4.6000}$^{16}$  & \colorbox{orange!38}{4.8300}$^{17}$  & \colorbox{orange!42}{30.6615}$^{16}$ & \colorbox{orange!14}{0.7751}$^{23}$ & \colorbox{orange!18}{-0.5028}$^{22}$ & \colorbox{orange!22}{20.7491}$^{21}$ \\
                Dreamlike Photoreal v2.0~\smallcite{dreamlike-photoreal-20}   & \colorbox{orange!38}{4.5600}$^{17}$  & \colorbox{orange!54}{4.9800}$^{13}$ & \colorbox{orange!76}{33.7712}$^{7~~}$ &  \colorbox{orange!60}{0.8344}$^{11}$ & \colorbox{orange!54}{-0.0859}$^{13}$ & \colorbox{orange!72}{21.4832}$^{8~~}$ \\
                Safe SD WEAK~\smallcite{SafeLatentDiffusion}                  & \colorbox{orange!34}{4.5300}$^{18}$  & \colorbox{orange!26}{4.7100}$^{20}$  & \colorbox{orange!34}{30.5644}$^{18}$ & \colorbox{orange!34}{0.8140}$^{18}$ & \colorbox{orange!34}{-0.2728}$^{18}$ & \colorbox{orange!42}{20.9899}$^{16}$ \\
                SD v1.4~\smallcite{Rombach_2022_sd15_sd21}                    & \colorbox{orange!30}{4.5200}$^{19}$  & \colorbox{orange!30}{4.7600}$^{19}$  & \colorbox{orange!26}{29.9149}$^{20}$ & \colorbox{orange!26}{0.8048}$^{20}$ & \colorbox{orange!30}{-0.3438}$^{19}$ & \colorbox{orange!30}{20.8462}$^{19}$ \\
                SD v1.5~\smallcite{Rombach_2022_sd15_sd21}                    & \colorbox{orange!26}{4.4500}$^{20}$  & \colorbox{orange!42}{4.9000}$^{16}$  & \colorbox{orange!30}{30.1673}$^{19}$ & \colorbox{orange!38}{0.8142}$^{17}$ & \colorbox{orange!38}{-0.2213}$^{17}$ & \colorbox{orange!34}{20.8640}$^{18}$ \\
                Safe SD MEDIUM~\smallcite{SafeLatentDiffusion}                & \colorbox{orange!22}{4.4000}$^{21}$  & \colorbox{orange!10}{4.5600}$^{24}$  & \colorbox{orange!46}{30.7820}$^{15}$ & \colorbox{orange!22}{0.7974}$^{21}$ & \colorbox{orange!26}{-0.3591}$^{20}$ & \colorbox{orange!46}{21.0257}$^{15}$ \\
                Redshift diffusion~\smallcite{redshiftdiffusion}              & \colorbox{orange!18}{4.3500}$^{22}$  & \colorbox{orange!22}{4.6700}$^{21}$  & \colorbox{orange!18}{29.2865}$^{22}$ & \colorbox{orange!30}{0.8066}$^{19}$ & \colorbox{orange!22}{-0.4172}$^{21}$ & \colorbox{orange!14}{20.6327}$^{23}$ \\
                Safe SD MAX~\smallcite{SafeLatentDiffusion}                   & \colorbox{orange!14}{4.3100}$^{23}$  & \colorbox{orange!14}{4.5900}$^{23}$ & \colorbox{orange!22}{29.8126}$^{21}$ & \colorbox{orange!10}{0.7601}$^{24}$ & \colorbox{orange!10}{-0.6095}$^{24}$ & \colorbox{orange!18}{20.7046}$^{22}$ \\
                Openjourney v2~\smallcite{openjourneV2}              & \colorbox{orange!10}{4.1500}$^{24}$  & \colorbox{orange!18}{4.6500}$^{22}$  & \colorbox{orange!14}{29.2389}$^{23}$ & \colorbox{orange!18}{0.7851}$^{22}$ & \colorbox{orange!14}{-0.6051}$^{23}$ & \colorbox{orange!10}{20.5973}$^{24}$ \\
                \bottomrule
          \end{tabular}
\end{table}

\mypara{Evaluation on image faithfulness.}
We evaluate image faithfulness on the testing dataset to ensure that the finetuned MLLM aligns with human judgment and generalizes to unseen data.
As detailed in Table~\ref{tab:human alignment faithfulness}, the finetuned MLLM successfully aligns with human preferences on image faithfulness, indicating its ability of image faithfulness evaluation is close to human.
Specifically, the rankings of the top and bottom 10 models by both \ourmethod and human evaluation scores are remarkably consistent.
Besides, most of the images in the testing dataset, especially those from the 16 exclusive generative models, are not present during the SFT training, showcasing the robust generalization capability of our models.

\mypara{Evaluation on text-image alignment.}
The evaluation of text-image alignment on the testing dataset is similar to that of image faithfulness.
Table~\ref{tab:human alignment faithfulness} reveals that the rankings of the 24 evaluated models by \ourmethod are generally consistent with human annotators.
We believe that the consistency on image faithfulness and text-image alignment evaluations mainly stems from our annotated high-quality SFT dataset.
It also proves that, with the annotated dataset and the extraordinary image-text joint understanding ability owned by MLLMs, we can easily finetune a MLLM to conduct the evaluation with low cost but close-to-human performance. 

\subsection{Ablations and Analyses of \ourmethod}

\mypara{Results on different prompt categories.}
Since MLLMs are not specifically trained to perform evaluations, they are naturally ill-suited for this task, hindering their task performances.
Therefore, we need to annotate SFT data for this task and finetune the MLLMs accordingly.
To verify the necessity, We conduct experiments comparing the LLava-Next 13B model with and without SFT. 
As shown in Table~\ref{tab:zero-shot-fidelity} and Table~\ref{tab:zero-shot-t2i}, the results demonstrate that SFT training considerably improves performance across all prompt categories in both image faithfulness and text-to-image alignment, closely aligning the MLLM's predictions with human evaluations.
Note that Table~\ref{tab:zero-shot-fidelity} illustrates that the baseline method without SFT performs poorly in image faithfulness and text-image alignment evaluations, particularly in the former.

\begin{table}

	\begin{minipage}[t]{.48\textwidth}
	\centering
        \small
        \tabcolsep 2.5pt
        \centering
        \fontsize{8}{8}\selectfont
        \caption{\small \textbf{Results of different prompt categories for evaluating image faithfulness.} Baseline is the vanilla LLaVA-NeXT model without find-tuning with human-aligned data.} \label{tab:zero-shot-fidelity}
            \begin{tabular}{l|ccccc}
            \toprule
            Method & Body & Hand & Face & Object & Common \\ 
            \midrule
            Human & 1.6701 & 1.0278 & 1.4107 & 2.2968  & 1.0637 \\ 
            Baseline & 3.9950 & 3.9932 & 3.9867 & 2.6734 & 3.3476 \\ 
            \ourmethod & 1.7305 & 0.9490 & 1.4393 & 2.3565 & 1.0903 \\ 
            \bottomrule
            \end{tabular}
	\end{minipage}
	\hspace{1mm}
        \begin{minipage}[t]{.5\textwidth}
	\centering
        \small
        \tabcolsep 1.8pt
        \fontsize{8}{8}\selectfont
        \caption{\small \textbf{Results of different prompt categories for evaluating text-to-image alignment.} Baseline is the vanilla LLaVA-NeXT model without find-tuning with human-aligned data.} 
        \label{tab:zero-shot-t2i}
            \begin{tabular}{l|cccccc}
            \toprule
            Method & Object & Count & Color & Style & Spatial & Action \\ 
            \midrule
                 Human & 1.6947 & 1.2032 & 1.8551 & 1.9796 & 1.5608 & 1.8015 \\ 
                 Baseline & 1.5602 & 1.0742 & 1.9275 & 1.1837 & 1.4118 & 1.1838 \\ 
                 \ourmethod & 1.6807 & 1.2516 & 1.8696 & 1.9592 & 1.5882 & 1.8382 \\ 
            \bottomrule
            \end{tabular}
	\end{minipage}
    \vspace{-2mm}
\end{table}

\begin{table}[t]
    \centering
    \small
    \tabcolsep 3pt 
    \caption{\small \textbf{Ablation study on the size of training data.} Results are reported on image faithfulness under different training data scale. We observe that a small number of annotated training data is sufficient for optimal results.}
    \label{tab:data_scale}
    \begin{tabular}{l|c|cccccccc}
            \toprule
             Method & Data Size                 & SDXL & Pixart & Wuerstchen & SDXL-Turbo & IF & SD v1.5 & SD v2.1 & LCM\\ 
             \midrule
             Human & --                         & 2.1044 & 1.8606 & 1.7839 & 1.3854 & 1.3822 & 1.3818 & 1.1766 & 1.0066 \\
             \midrule
             \multirow{3}{*}{\ourmethod}  & 200 & 1.7443 & 1.8898 & 1.9278 & 1.1261 & 1.2977 & 1.5254 & 1.4309 & 1.1204 \\ 
              & 500                             & 1.8890 & 1.9161 & 1.8586 & 1.2141 & 1.3109 & 1.3926 & 1.3815 & 0.9485 \\ 
              & 800                             & 2.0443 & 1.9199 & 1.8012 & 1.3353 & 1.296 & 1.4702 & 1.3221 & 1.0305 \\ 
            \bottomrule
          \end{tabular}
\end{table}
\begin{table}[t]
    \centering
    \small
    \tabcolsep 3pt 
    \caption{\small \textbf{Ablation study on the size vision-language model.} Results are reported on image faithfulness under different model scales of LLaVA-NeXT. We observe that model size is critical for reliable evaluation.}
    \vspace{3mm}
    \label{tab:model_scale}
      \begin{tabular}{l|c|cccccccc}
        \toprule
         Method & Model Size & SDXL & Pixart & Wuerstchen & SDXL-Turbo & IF & SD v1.5 & SD v2.1 & LCM \\ 
         \midrule
         Human & -- & 2.1044 & 1.8606 & 1.7839 & 1.3854 & 1.3822 & 1.3818 & 1.1766 & 1.0066 \\ 
         \midrule
         \multirow{3}{*}{\ourmethod} & 7B & 1.9959 & 1.8615  & 1.8228 & 1.1708  & 1.2704 & 1.4031  & 1.3063       & 1.0145  \\ 
         & 13B  & 2.0443 & 1.9199 & 1.8012 & 1.3353 & 1.2960  & 1.4702  & 1.3221        & 1.0305 \\
         & 34B  & 2.1131 & 1.8621 & 1.8083 & 1.3906 & 1.3076  & 1.3921  & 1.2037        & 1.0143 \\ 
        \bottomrule
      \end{tabular}
\end{table}

\mypara{Effect of training dataset size for vision-language model training.}
In order to explore the effects of data size and determine the sufficient amount of training data, we train the model on image faithfulness evaluation task with images and their annotations sourced from 200, 500 and 800 prompts. 
As illustrated in Table~\ref{tab:data_scale}, the evaluation performance continuously enhances as more training data is used.
Notably, training with just 500 prompts nearly maximizes accuracy, with further increases to 800 data yielding only marginal improvements.
This result suggests that our method requires only a small amount of annotated data to achieve good performance, highlighting its cost-effectiveness.
Generally, since more data leads to better performance, we use all of the available data to finetune our models and release this data to the research community to bootstrap further study.

\mypara{Effect of model size.}
Since transformers are known for their scalability~\citep{radford2018improving, dehghani2023scaling}, we investigate the effect of the model size on the performance of image faithfulness evaluation. 
As illustrated in Table~\ref{tab:model_scale}, the benefits of scaling up the utilized MLLMs are remarkably significant, where increasing the model size from 7B to 34B results in substantial improvements in evaluation performance.
For this consequence, for the final version of the \ourmethod evaluation model, we choose LLaVA-NeXT 34B, the largest model in LLaVA-NExT series, and finetune it on our meticulously curated SFT data.
Since some users of \ourmethod cannot afford MLLM inference with 34B parameters, we will make the 13B and 34B models publicly available.

\subsection{Comparison with Existing Evaluation Methods}

\mypara{SFT with human-aligned data outperforms vanilla MLLMs.}
To validate the effectiveness of the MLLM after SFT, we use vanilla LLaVA-NeXT 13B as the baseline model for comparison. 
As shown in Table~\ref{tab:zero-shot-fidelity} and Table~\ref{tab:zero-shot-t2i}, the results of vanilla model suggest some correlations with human-annotated data. 
However, the alignment of the vanilla MLLM is relatively low due to the absence of images generated by model (such as distorted bodies and hands images) and issues related to evaluation in the MLLM's pre-training dataset.
After applying SFT on the LLaVA-Next 13B model using human annotated data, the model's predictions on various fine-grained evaluation metrics are almost align to the human-annotated data and significantly surpass the evaluation results of all MLLM models that are not finetuned. 
This experimental results confirms that our SFT training enables the MLLM to be successfully applied to the task of evaluating text-to-image models.

\mypara{Comparison with other methods.}
To verify the human preference alignment of our model, especially when compared with other baseline methods, we calculate Kendall rank~\citep{KENDALL1938} and Pearson~\citep{freedman2007statistics} correlation coefficient on images generated by 24 text-to-image models and summarize the results in Table~\ref{tab:comparison}.

\begin{wraptable}{r}{0.5\linewidth}
    \vspace{-8mm}
    \centering
    \small
    \tabcolsep 2pt 
    \caption{\small \textbf{Comparison with existing methods.}}\label{tab:comparison}
    \vspace{2mm}
    \begin{tabular}{l|cc|cc}
    \toprule
    \multirow{2.5}{*}{Method} & \multicolumn{2}{c|}{Faithfulness} & \multicolumn{2}{c}{Alignment} \\
    \cmidrule{2-5}
    & Kendall$\uparrow$  & Pearson$\uparrow$ & Kendall$\uparrow$ & Pearson$\uparrow$ \\
    \midrule
     CLIP-score & 0.1304 & 0.1765 & 0.6956 & 0.8800 \\
     HPSv2  & 0.4203 & 0.5626 & 0.5217 & 0.7113 \\
     \ourmethod   & 0.7464 & 0.8730 & 0.8043 & 0.9356 \\
     \bottomrule
    \end{tabular}
\end{wraptable}
As can be concluded, compared with baseline methods, \ourmethod achieves significant higher alignment with fine-grained human preference on image faithfulness and image-text consistency, showcasing robust generalization ability.
Although HPS v2 roughly aligns with human preference in some extent, the relative small model capacity and coarse ranking training limits its generalization to the fine-grained annotated data.
Besides, since CLIP-s only cares the CLIP similarity of the generated image and its corresponding prompt, it behaves poorly in image faithfulness evaluation. The per-question alignment and the leaderboard of \ourmethod will be introduced in the supplementary materials.

\section{Conclusion and Discussion}
\label{sec:disscusion}

In this work, we design an economic evaluation method that offers high accuracy, strong generalization capabilities, and provides fine-grained, interpretable metrics. 
We develop a comprehensive data annotation and cleaning process tailored for evaluation tasks, and establish the \ourmethod benchmark for training and evaluating models on supervised fine-tuning tasks for MLLMs. 
Experimental results across 24 text-to-image models demonstrate that our evaluation metrics surpass the accuracy of all the state-of-art evaluation method. 
Additionally, we conduct a detailed empirical study on how MLLMs can be applied to model evaluation tasks.
There are still many opportunities for further advancements and expansions based on our \ourmethod. We hope that our work can inspire and facilitate future research in this field.

\section{Reproducibility Statement}
The full version of the source code, dataset, as well as the final version of the finetuned MLLMs~(one finetuned on LLaVA-NeXT 13B and the other one finetuned on LLaVA-NeXT 34B) will be released to the public. 
The data construction procuedure, including data collection and curation, data cleaning and annotation, is thoroughly described in Section~3.
For details related to the human annotation and the measures that used to ensure its quality, we comprehensively introduce them in Appendix~B.
As for every experiment introduced in this paper, we provide a general introduction in Section~\ref{sec:experiments} and exhibit implementation details related to reproduce our experiments.
Specifically, the latter includes the hyper-parameters of each evaluated models, the employed instruction, as well as more supplementary experiments, which are described in Appendix~C, Appendix~D and Appendix~E.

\section{Ethics Statement}
We are committed to conducting this research with the highest ethical standards. 
Our goal is to contribute positively to the fields of evaluation benchmarks on artificial intelligence generated content, emphasizing transparency and reproducibility in our design.
Similar with other MLLMs, \ourmethod may potentially generate responses contain offensive, inappropriate, or harmful content.
Since the base MLLMs of \ourmethod are pretrained on large datasets scraped from the web that might contain private information and harmful content, they may inadvertently generate or expose sensitive information, raising ethical and privacy concerns.
MLLMs are also susceptible to adversarial attacks, where inputs are intentionally crafted to deceive the model.
This vulnerability can be exploited to manipulate model outputs, posing security and ethic risks.
To alleviate these safety limitation and our fulfill our social responsibility as artificial intelligence researchers, we create dedicated evaluation sets for bias detection and mitigation, and conducted adversarial testing through hours of redteaming.
Besides, \ourmethod is designed for fine-grained, human-aligned automatic text-to-image evaluations, which can serve as a stepping stones toward revealing the inner generation nature of text-to-image generative models, thereby lowering the ethical hazard of these models.
We believe that with appropriate use, it could provide users with interesting experiences for detailed synthesized image evaluation, and inspires more appealing research works about text-to-image generation.

\small{
\bibliography{main}
\bibliographystyle{iclr2025_conference}
}
\appendix
\section{Limitations}
\label{supp_sec:limitation}

\mypara{Multimodal LLMs.}
Since \ourmethod evaluation models are fine-tuned MLLMs, they also suffer from multimodal hallucination, where models may generate content that seems plausible but actually incorrect or fabricated, and cannot be inferred from the input images and texts.
Moreover, due to the possible harmful content in the pretraining data of the utilized base MLLMs, the model may inherit these biases and generate inappropriate response.
Although we carefully curate the SFT training data of the \ourmethod evaluation models, the problems of hallucination and biased pre-training is alleviated but not fully addressed.
Other than the these issues, \ourmethod evaluation models also suffer from opacity and interpretability, context limitation, as well as sensitivity to input formatting, like most multimodal LLMs.

\mypara{Human Annotations.}
Human annotation is naturally subjective and influenced by individual perspectives, biases, and preferences.
During the annotation, annotators can make mistakes, leading to incorrect or noisy labels. 
Regarding these challenges, we conduct 9 rounds of trial annotation and 2 rounds of random sampling quality inspection to ensure the inter-annotator consistency and overall annotation quality.
We also design easy-to-understand annotation guidelines, instructions and platform to lower the annotation difficulty and benefit the annotation accuracy.
Despite all these efforts, conducting human annotation with different annotators, user interface and annotation guidelines may lead to different result, making our annotation somewhat limited.
Furthermore, human annotation can be time-consuming and resource-intensive, limiting the scale at which we can afford.

\section{Annotation Details}
\label{supp_sec:annotation_detail}
Before performing the final human annotation, we made a series of efforts to guarantee its quantity, quality and efficiency.
To begin with, we select appropriate candidates to perform the annotation and hold a training meeting for them.
Then, we design a user-friendly user interface and a comprehensive annotation procedure.
We write a detailed annotation guidelines to explain every aspect and precaution of the annotation.
As mentioned above, we conduct 9 rounds of trial annotation on another 50 synthesized images and 2 turns of random sampling quality inspection to further ensure inter-annotator consistency and annotation accuracy.

\mypara{Annotator selection.}
The accuracy and reliability of the annotated data depend heavily on the capabilities of the human annotators involved in the annotation process.
As a consequence, at the beginning of the annotation, We first conduct annotator selection to build an appropriate and unbiased annotation team, and train this annotation team with our meticulously prepared annotation guidelines.
For annotator selection, we let the candidates to accomplish a test concentrating on 10 factors, domain expertise, resistance to visually disturbing content, attention to detail, communication skills, reliability, cultural and linguistic competence, technical skills, ethical considerations, aesthetic cognition, and motivation.
Notably, since the evaluated models may generate images with uncomfortable and inappropriate visual content, the candidates are notified with this inconvenience before the test.
Only those agreed with this inconvenience are eligible to participate in the test, and they are welcome to withdraw at any time if they choose to do so.
Based on the test results and candidate backgrounds, We try our best to ensure that the selected annotators are well-balanced in background and have a generally competitive abilities of the 10 mentioned factors.
To summarize, our annotation team includes 10 annotators carefully selected from 29 candidates, 5 males and 5 females, all have a bachelor's degree. 
We interview the annotators and ensure they are adequate for the annotation.

\begin{figure}[t]
    \centerline{\includegraphics[width=0.95\linewidth]{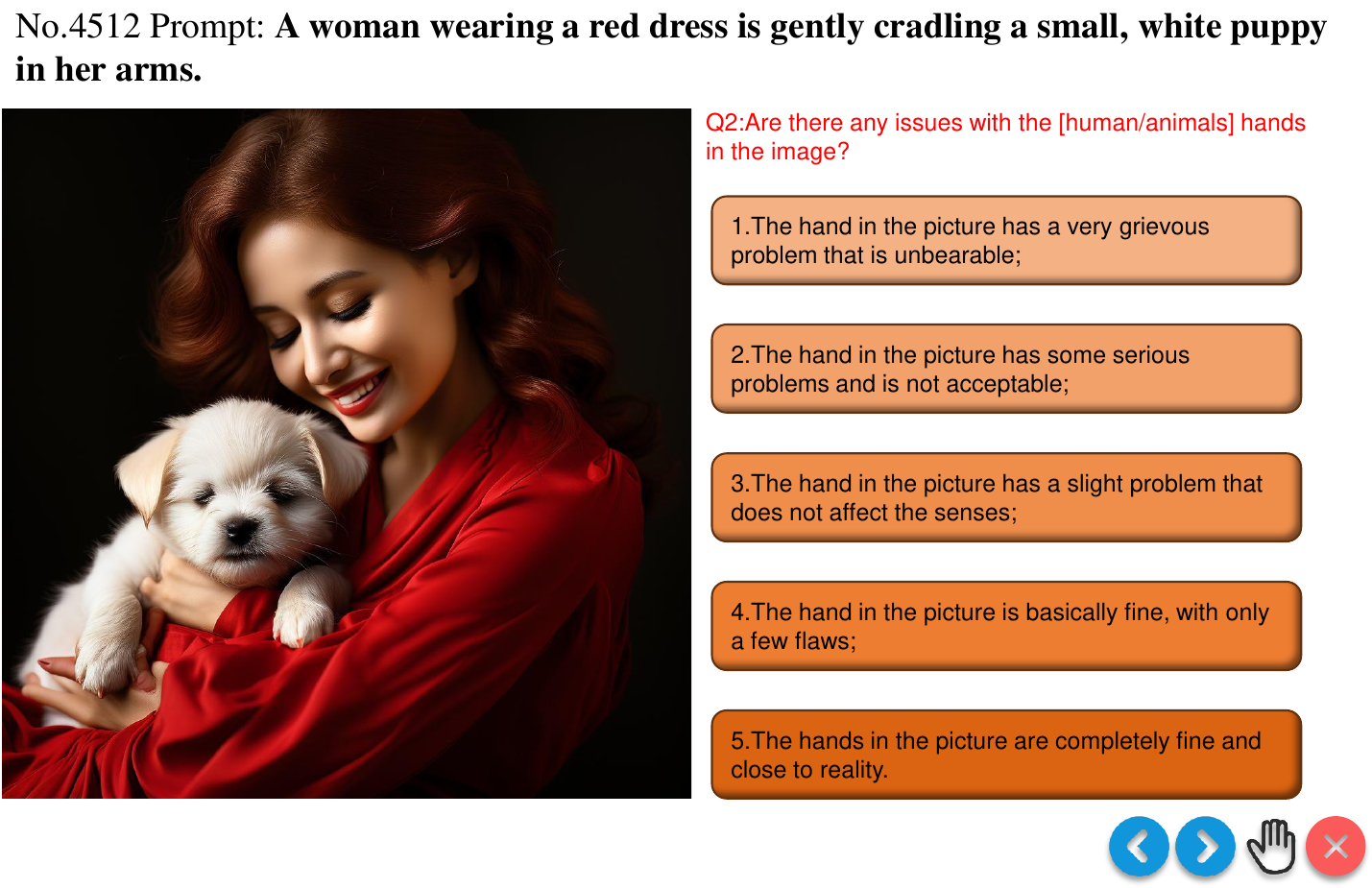}}
    \vspace{-4mm}
    \caption{\small Demonstration of our user interface.
    Each time, our specially designed user interface will provide one sample to the annotators.
    We incorporated four distinct icons to signify various functionalities of the user interface.
    }
    \vspace{-12pt}
    \label{fig:user_interface}
\end{figure}

\mypara{Annotation training and guidelines.}
After the selection, we conduct a training meeting over our comprehensive user guidelines to make the annotation team aware of our purpose and standard.
During the training meeting, we explain the purpose, precaution, standard, workload and wage of the annotation.
Besides, we have formally informed the annotators that the human annotation is solely for research purposes, and the data they have annotated may potentially be released to the public in the future.
We, and the annotators reached consensus on the standard, workload, wage and intended usage of the annotated data.
The rules for recognising image faithfulness and text-image are universal, and thus each individual's standards should not differ significantly.
As a consequence, we annotate a few samples using our meticulously developed annotation platform for the annotators to ensure inter-annotator consistency.
The overall snapshot of the developed annotation paltform is exhibited in~\cref{fig:user_interface}. 
With this training, we also equip the annotators with necessary knowledge for unbiased detailed human evaluation on image faithfulness and text-image alignment.
Specifically, the employed annotation guidelines involve the instructions for using the annotation platform and detailed guidelines about the annotation procedure, and we demonstrate them in Table~\ref{tab:user_guidelines}.
\begin{table}[h]
\footnotesize
\centering
\small
\caption{\small
User Guidelines of the Human Annotation.
Considering that our annotators are native Chinese speakers while our readers may not be, each user is actually provided with a copy of Chinese version of the user guidelines. 
Meanwhile, we demonstrate its translated English version as follows.
}
\begin{tabular}{|p{1\columnwidth}|}
\bottomrule
\textbf{User Guidelines}
\\ \hline
\textbf{Part I Introduction}

Welcome to the annotation platform. This platform is designed to simplify the annotation process and enhance annotation efficiency. Before the detailed introduction, we want to claim again that you may feel inconvenient as the evaluated models may generate images with uncomfortable and inappropriate visual content. Now, you are still welcomed if you want to withdraw your consent
The annotation process is conducted on a sample-by-sample basis, with a question-by-question approach. Thus, you are supposed to answer all the questions raised for the present sample to accomplish its annotation. Once all the delegated samples are accomplished, your job is finished and we are thankful for your contribution to the project.
\\ \hline
\textbf{Part II Guidelines of the User Interface} 

1.User Login: To access the annotation platform, you are required to login as a user. Please navigate to the login page, enter the username and password provided by us, and click the ``Login'' button.

2.Dashboard: Once you complete the login, you will be jumped into the dashborad page. The dashboard will list the overview of the samples assigned to you to annotate. Besides, we list the status of each sample for you to freely check your annotation progress (e.g., pending, completed). 

3.Annotation Interface: Click on the ``Start'' button or an assigned image through the dashboard interface, you will jump into the annotation interface. annotation interface is made up of three components: 1) Image Display: View the image to be annotated and its conditioned prompts; 2) Question Panel: List of single-choice questions related to the image; 3) Navigation Buttons: "Next" and "Previous" buttons to navigate through questions and images.

4.Answering Questions: Each time, the annotation interface will provide you a sample for annotation, please view the image and read the associated question, select the appropriate answer from the available options, and repeat the processs for all questions related to the question.

5.Saving and Submitting Annotations: To save progress and submit completed annotations, you can click the ``Save'' button to save your progress. If you finish the assigned sample and ensure the accuracy and confidence of its, you can click the ``Submit'' button to submit this annotation.

6.Review and Edit Annotations: If you want to review and edit your submission, you can navigate the completed tasks section, and select the image to review. You will jump into its annotation interface with the previously submitted annotations and are allowed to do any modification.

7.Report and contact: If you find any problem about the assigned sample, such as witnessing NSFW or biased content, assigned visually abnormal sample, feel free to click the ``Report'' button and fill a form to report this sample. If you have any question about the standard of the annotation or have suggestions for improvement, please do not hesitate to contact us through phone, we will be glad to help you.

\\ \hline
\textbf{Part III General Guidelines of the Human Annotation}

1.In general, you are supposed to answer all the questions raised for the present sample to accomplish its annotation. This annotation only involves single-choice question.

2.Before answering the question, please ensure that the question is applicable to this prompt. If it is not applicable, please select option 0 directly—this is the predefined option for this particular scenario.

3.If you are answering a question about the image faithfulness, you may find the question is applicable to multiple objects within the image. you need to answer the question regarding to every applicable object and its role in the image. A straightforward way for this is to solely score every applicable object and choose the option closest to the calculated weighted average score.

4.If you are answering the object faithfulness question on the image faithfulness annotation, you need to drop and report for the encountered image with no clear main object. 

5.If you are answering the commonsense question on the image faithfulness annotation, you need to drop and report for the encountered surreal and sci-fi image.

6.You are required to first annotate 30 samples to form a stable and reasonable assessment standard. Then, accomplish the annotation in progress.

7.This annotation is for evaluating image faithfulness and text-image alignment, as a consequence, the standard of the annotation is universal.

8.If you feel confused at anything about the human annotation, feel free to contact us through phone, we will be glad to help you.

9. Once you have submitted your annotation results, we are very thankful to inform you that you have finished your job. Thank you once again for your contribution to our project.

10.If you have submitted your annotation but want to withdraw your submission and review the annotation results, you can contact us through phone, we will send it back to you.
\\ \hline
\end{tabular}
\label{tab:user_guidelines}
\end{table}

\mypara{Trial Annotation}
Even with the above preparation, there is no quantitative evidence to verify the quality,  the efficiency, and the inter-annotator consistency of the human annotation.
Additionally, the standard for assessing image faithfulness and text-image are universal, which further emphasize the significant role of high inter-annotator consistency.
Considering that, we conduct a multi-turn trial annotation on another 50 synthesized images.
After each trial, we calculate the Cohen's kappa coefficient and conduct a meeting for our annotators to explain annotation standards, rules and guidelines, thereby ensuring high inter-annotator reliability.
In total, we conduct nine turns of trial annotation, and in the last turn of the trial, the Cohen's kappa coefficient of our annotators reaches $0.681$, indicating high inter-annotator reliability.

\mypara{Random Sampling Quality Inspection}
Upon reaching the milestone percentages of 25\%, 50\%, 75\%, and 100\% in the annotation progress, we conducted a series of random sampling quality inspections on the present annotation results at each milestone, totally four turns of random sampling quality inspection.
The random sampling quality inspection by four experts in text-to-image generation selected from our group on 1,000 randomly sampled annotated images.
For the first two turn of quality inspection, there are totally 423 and 112 annotated samples that failed the inspection.
The failed samples are re-annotated and re-inspected.
As for the last two turns of quality inspection, they both revealed zero failed samples due to the thoughtful and rigorous annotation preparation.

\section{Additional Details of the Evaluated Models}
\label{supp_sec:inference_setting}
In this section, we introduce the details of the evaluated text-to-image generative models in this work.

\begin{itemize}[itemsep=1pt,topsep=0.5pt,leftmargin=12pt]

\item \mypara{Stable Diffusion \{v1.4, v1.5, v2 base, v2.0, v2.1\}.} 
Stable Diffusion~(SD) is a series of 1B text-to-image generative models based on latent diffusion model~\citep{rombach2022latentdiffusionmodel} and is trained on LAION-5B~\citep{schuhmann2022laion}. Specifically, the SD series includes SD v1.1, SD v1.2, SD v1.4, SD v1.5, SD v2 base, SD v2.0, and SD v2.1 respectively. Among them, we choose the most commonly-employed SD v1.4, SD v1.5, SD v2.0 and SD v2.1 for \ourmethod evaluation.

SD v1.1 was trained at a resolution of 256x256 on laion2B-en for 237k steps, followed by training at a resolution of 512x512 on laion-high-resolution~((170M examples from LAION-5B with resolution >= 1024x1024) for the subsequent 194k steps. While, SD v1.2 was initialized from v1.1 and further finetuned for 515k steps at resolution 512x512 on laion-aesthetics v2 5+~(a subset of laion2B-en, filtered to images with an original size >= 512x512, estimated aesthetics score > 5.0, and an estimated watermark probability < 0.5). \textbf{SD v1.4} is initialized from v1.2 and subsequently finetuned for 225k steps at resolution 512x512 on laion-aesthetics v2 5+. This version incorporates a 10\% dropping of the text-conditioning to improve classifier-free guidance sampling. Similar to SD v1.4, \textbf{SD v1.5} is resumed from SD v1.2 and trained 595k steps at resolution 512x512 on laion-aesthetics v2 5+, with 10\% dropping of the text-conditioning.
    
SD v2 base is trained from scratch for 550k steps at resolution 256x256 on a subset of LAION-5B filtered for explicit pornographic material, using the LAION-NSFW classifier with punsafe = 0.1 and an aesthetic score >= 4.5. Then it is further trained for 850k steps at resolution 512x512 on the same dataset on images with resolution >= 512x512. \textbf{SD v2.0} is resumed from stable-diffusion v2 base and trained for 150k steps using a v-objective on the same dataset. After that, it is further finetuned for another 140k steps on 768x768 images. \textbf{SD v2.1} is finetuned from SD v2.0 with an additional 55k steps on the same dataset (with punsafe=0.1), and then finetuned for another 155k extra steps with punsafe=0.98.

\item \mypara{Stable Diffusion XL \{v1.0, Refiner v1.0\}.}
Stable Diffusion XL~(SDXL) is a powerful text-to-image generation model that iterates on the previous Stable Diffusion models in three key ways:
(1) its UNet is 3x larger and SDXL combines a second text encoder (OpenCLIP ViT-bigG/14) with the original text encoder to significantly increase the number of parameters;
(2) it introduces size and crop-conditioning to preserve training data from being discarded and gain more control over how a generated image should be cropped; 
(3) it introduces a two-stage model process; the base model (can also be run as a standalone model) generates an image as an input to the refiner model which adds additional high-quality details.

\item \mypara{Pixart-Alpha.}
Pixart-Alpha is a model that can be used to generate and modify images based on text prompts. 
It is a Transformer Latent Diffusion Model that uses one fixed, pretrained text encoders (T5)) and one latent feature encoder (VAE).

\item \mypara{Latent Consistency Model Stable Diffusion XL}
Latent Consistency Model Stable Diffusion XL~(LCM SDXL) \cite{luo2023lcm} enables SDXL for swift inference with minimal steps.
Viewing the guided reverse diffusion process as solving an augmented probability flow ODE (PF-ODE), LCMs are designed to directly predict the solution of such ODE in latent space, mitigating the need for numerous iterations and allowing rapid, high-fidelity sampling.

\item \mypara{Dreamlike Diffusion 1.0.} 
Dreamlike Diffusion 1.0 \citep{dreamlike-diffusion-10} is a SD v1.5 model finetuned on high-quality art images by dreamlike.art.

\item \mypara{Dreamlike Photoreal 2.0.} 
Dreamlike Photoreal 2.0 \citep{dreamlike-photoreal-20} is a photorealistic text-to-image latent diffusion model resumed from SD v1.5 by dreamlike art.
This model was finetuned on 768x768 images, it works pretty good with resolution 768x768, 640x896, 896x640 and higher resolution such as 768x1024.

\item \mypara{Openjourney v1, v2.}
Openjourney \citep{openjourne} is an open-source text-to-image generation model resumed from SD v1.5 and finetuned on Midjourney images by PromptHero. 
Openjourney v2 \citep{openjourneV2} was further finetuned using another 124000 images for 12400 steps, about 4 epochs and 32 training hours.

\item \mypara{Redshift Diffusion.} 
Redshift Diffusion \citep{redshiftdiffusion} is a Stable Diffusion model finetuned on high-resolution 3D artworks. 

\item \mypara{Vintedois Diffusion.}
Vintedois Diffusion \citep{vintedoisdiffusion} is a Stable Diffusion v1.5 model finetuned on a large number of high-quality images with simple prompts to generate beautiful images without a lot of prompt engineering. 

\item \mypara{Safe Stable Diffusion \{Weak, Medium, Strong, Max\}.}
Safe Stable Diffusion \citep{SafeLatentDiffusion} is an enhanced version of the SD v1.5 model by mitigating inappropriate degeneration caused by pretraining on unfiltered web-crawled datasets.
For instance SD may unexpectedly generate nudity, violence, images depicting self-harm, and otherwise offensive content.
Safe Stable Diffusion is an extension of Stable Diffusion that drastically reduces this type of content. 
Specifically, it has an additional safety guidance mechanism that aims to suppress and remove inappropriate content (hate, harassment, violence, self-harm, sexual content, shocking images, and illegal activity) during image generation. 
The strength levels for inappropriate content removal are categorized as: \{Weak, Medium, Strong, Max\}.

\item \mypara{MultiFusion.}
MultiFusion \citep{MultiFusion} is a multimodal, multilingual diffusion model that extends the capabilities of SD v1.4 by integrating various modules to transfer capabilities to the downstream model.
This combination results in novel decoder embeddings, which enable prompting of the image generation model with interleaved multimodal, multilingual inputs, despite being trained solely on monomodal data in a single language.

\item \mypara{DeepFloyd-IF \{ M, L, XL \} v1.0.} 
DeepFloyd-IF \citep{DeepFloydIF} is a novel state-of-the-art open-source text-to-image model with a high degree of photorealism and language understanding.
It is a modular composed of a frozen text encoder and three cascaded pixel diffusion modules: a base model that generates 64x64 image based on text prompt and two super-resolution models, each designed to generate images of increasing resolution: 256x256 and 1024x1024, respectively. 
All stages of the model utilize a frozen text encoder based on the T5 transformer to extract text embeddings, which are then fed into a UNet architecture enhanced with cross-attention and attention pooling.
Besides, it underscores the potential of larger UNet architectures in the first stage of cascaded diffusion models and depicts a promising future for text-to-image synthesis.
The model is available in three different sizes: M, L, and XL. M has 0.4B parameters, L has 0.9B parameters, and XL has 4.3B parameters.

\end{itemize}


\clearpage
\section{Instruction Templates}
\label{supp_sec:instruction_templates}
Here, we present every instruction used for \ourmethod evaluation on image faithfuleness and text-image alignment.
The templates contain some placeholders set for filling in the corresponding attributes of the input images during the evaluation.
For example, a specific ``<ObjectHere>'' and ``<NumberHere>'' can be ``people, laptop, scissors.'' and ``plate: 1,  turkey sandwich: 3, lettuce: 1.'', respectively.

For \ourmethod evaluation on image faithfulness, we devise 5 questions concentrate on the faithfulness of the generated body structure, generated face, generated hand, generated objects, as well as generation adherence to commonsense and logic.
The instruction templates for these fine-grained criteria are as follows:

\begin{tcolorbox}
\textbf{[Q1]}:Are there any issues with the [human/animals] body structure in the image, such as multiple arms, missing limbs or legs when not obscured, multiple heads, limb amputations, and etc?

\textbf{[OPTIONS]}:
0.There are no human or animal body in the picture;
1.The body structure of the people or animals in the picture has a very grievous problem that is unbearable;
2.The body structure of the people or animals in the picture has some serious problems and is not acceptable;
3.The body structure of the people or animals in the picture has a slight problem that does not affect the senses;
4.The body structure of the people or animals in the picture is basically fine, with only a few flaws;
5.The body structure of the people or animals in the picture is completely fine and close to reality.

\end{tcolorbox}
\begin{tcolorbox}
\textbf{[Q2]}:Are there any issues with the [human/animals] hands in the image, such as having more or less than five fingers when not obscured, broken fingers, disproportionate finger sizes, abnormal nail size proportions, and etc?

\textbf{[OPTIONS]}:
0.No human or animal hands are shown in the picture;
1.The hand in the picture has a very grievous problem that is unbearable; 
2.The hand in the picture has some serious problems and is not acceptable; 
3.The hand in the picture has a slight problem that does not affect the senses;
4.The hand in the picture is basically fine, with only a few flaws;
5.The hands in the picture are completely fine and close to reality.

\end{tcolorbox}
\begin{tcolorbox}
\textbf{[Q3]}:Are there any issues with [human/animals] face in the image, such as facial distortion, asymmetrical faces, abnormal facial features, unusual expressions in the eyes, and etc?

\textbf{[OPTIONS]}:
0.There is no face of any person or animal in the picture;
1.The face of the person or animal in the picture has a very grievous problem that is unbearable;
2.The face of the person or animal in the picture has some serious problems and is not acceptable;
3.The face of the person or animal in the picture has a slight problem that does not affect the senses;
4.The face of the person or animal in the picture is basically fine, with only a few flaws;
5.The face of the person or animal in the picture is completely fine and close to reality.

\end{tcolorbox}
\begin{tcolorbox}
\textbf{[Q4]}:Are there any issues or tentative errors with objects in the image that do not correspond with the real world, such as distortion of items, and etc?

\textbf{[OPTIONS]}:
0.There are objects in the image that completely do not match the real world, which is very serious and intolerable;
1.There are objects in the image that do not match the real world, which is quite serious and unacceptable;
2.There are slightly unrealistic objects in the image that do not affect the senses;
3.There are basically no objects in the image that do not match the real world, only some flaws;
4.All objects in the image match the real world, no problem.

\end{tcolorbox}
\begin{tcolorbox}
\textbf{[Q5]}:Does the generated image contain elements that violate common sense or logical rules, such as animal/human with inconsistent anatomy, object-context mismatch, impossible physics, scale and proportion issues, temporal and spatial inconsistencies, hybrid objects, and etc?

\textbf{[OPTIONS]}:
0.The image contains elements that violate common sense or logical rules, which is very grievous and intolerable;
1.The presence of elements in the image that seriously violate common sense or logical rules is unacceptable;
2.The image contains elements that violate common sense or logical rules, which is slightly problematic and does not affect the senses;
3.There are basically no elements in the image that violate common sense or logical rules, only some flaws;
4.There are no elements in the image that violate common sense or logical rules, and they are close to reality.
\end{tcolorbox}

The templates of \ourmethod evaluation on text-image alignment are as follows.
We select 6 common aspects of text-image alignment, object, number, color, style, spatial relationship and action. 
For images that do not involve the specified attribute, the corresponding question template is not filled in and subsequently input into \ourmethod.

\begin{tcolorbox}
\textbf{[Q1]}:Does the given image contain all the objects (<ObjectHere>) presented in the corresponding prompts?

\textbf{[OPTIONS]}:
1.None objects are included;
2.Some objects are missing;
3.All objects are included.
\end{tcolorbox}

\begin{tcolorbox}
\textbf{[Q2]}:Does the given image correctly reflect the numbers (<NumberHere>) of each object presented in the corresponding prompts? 

\textbf{[OPTIONS]}:
1.All counting numbers are wrong;
2.Some of them are wrong;
3.All counting numbers are right.
\end{tcolorbox}

\begin{tcolorbox}
\textbf{[Q3]}:Does the given image correctly reflect the colors of each object (<ColorHere>) presented in the corresponding prompts?

\textbf{[OPTIONS]}:
1.All colors are wrong;
2.Some of them are wrong;
3.All corresponding colors numbers are right.
\end{tcolorbox}

\begin{tcolorbox}
\textbf{[Q4]}:Does the given image correctly reflect the style (<StyleHere>) described in the corresponding prompts? 

\textbf{[OPTIONS]}:
1.All styles are wrong;
2.Some of them are wrong;
3.All styles are right.
\end{tcolorbox}

\begin{tcolorbox}
\textbf{[Q5]}:Does the given image correctly reflect the spatial relationship (<SpatialHere>) of each object described in the corresponding prompts?

\textbf{[OPTIONS]}:
1.All spatial relationships are wrong;
2.Some of them are wrong;
3.All spatial relationships are right.
\end{tcolorbox}

\begin{tcolorbox}
\textbf{[Q6]}:Does the given image correctly reflect the action of each object (<ActionHere>) described in the corresponding prompts?

\textbf{[OPTIONS]}:
1.All actions are wrong;
2.Some of them are wrong;
3.All actions are right.
\end{tcolorbox}

\clearpage
\section{Additional Quantitative Analysis}
\label{supp_sec:supp_exp}
\subsection{Generalization Experiments}
To verify the generalization capability of our evaluation model, We compared MLLM's SFT using different training datasets: one with images generated by all 8 text-to-image models and another with images generated by only 4 of these models, while the final evaluation was conducted on images generated by the other 4 models.
As shown in Table~\ref{tab:generalization_faithfulness} and Table~\ref{tab:generalization_alignment},
We observed that MLLMs trained on images from a subset of text-to-image models can effectively generalize to images generated by unseen text-to-image models.

\begin{table}[h]
    \centering
    \small
    \tabcolsep 8pt 
    \caption{\textbf{Ablation study on the number of different text-to-image models used to generate the training data for evaluating image faithfulness.} 
                     We observe that \ourmethod exhibits strong generalization capability. }
    \label{tab:generalization_faithfulness}
    \begin{tabular}{l|c|cccccc}
        \toprule
         Method  &  T2I models  & body & hand & face & object & common & MAE\\ 
         \midrule
         Human      &  -  & 1.4988 & 0.8638 & 1.1648 & 2.2096  & 0.8710 & 0\\ 
         \midrule
         \multirow{2}{*}{\ourmethod} &  8  & 1.6058 & 0.7901 & 1.1974 & 2.2783 & 0.8871 & 0.0596\\ 
                    &  4  & 1.6522 & 0.9588 & 1.2355 & 2.3032 & 0.9516 & 0.0987\\ 
        \bottomrule
    \end{tabular}
\end{table}
\begin{table}[h]
    \centering
    \small
    \tabcolsep 7pt 
    \caption{\textbf{Ablation study on the number of different text-to-image models used to generate the
training data for evaluating text-to-image alignment.} We observe that EVALALIGN exhibits strong
generalization capability.}
    \label{tab:generalization_alignment}
    \begin{tabular}{l|c|ccccccc}
        \toprule
         Method      &T2I models& Object & Count & Color & Style & Spatial & Action & MAE\\ 
         \midrule
         Human       & - & 1.7373 & 1.3131 & 2.0000 & 1.9333 & 1.5952 & 1.8837 & 0 \\ 
         \midrule
         \multirow{2}{*}{\ourmethod}    & 8 & 1.7203 & 1.3232 & 1.9565 & 1.9333 & 1.6547 & 1.8605 & 0.0256\\ 
                 & 4 & 1.7832 & 1.3526 & 1.9637 & 1.9876 & 1.6891 & 1.8954 & 0.0469\\ 
        \bottomrule
    \end{tabular}
\end{table}

\subsection{Instruction Enhancement Experiments}

\begin{table}[t]
    \centering
    \small
    \tabcolsep 1.5pt 
    \caption{\small \textbf{Ablation study on the enhancement of instructions.} 
    Results are reported on image faithfulness under different instructions. 
    We observe that enhanced instructions can significantly improves the evaluation metrics. MAE: mean absolute error.}
    \label{tab:instruction_enhancement}
    \begin{tabular}{l|c|cccccccc|c}
            \toprule
             Method & Instruction                 & SDXL & Pixart & Wuerstchen & SDXL-Turbo & IF & SD v1.5 & SD v2.1 & LCM &MAE\\ 
             \midrule
             Human & --                          & 2.1044 & 1.8606 & 1.7839 & 1.3854 & 1.3822 & 1.3818 & 1.1766 & 1.0066 & 0\\
             \midrule
             \multirow{3}{*}{\ourmethod}  & \xmark  & 1.9565 & 1.9286 & 1.8565 & 1.1818 & 1.3419 & 1.4801 & 1.4078 & 1.1051 & 0.1201 \\ 
              & \cmark                               & 2.0443 & 1.9199 & 1.8012 & 1.3353 & 1.2960 & 1.4702 & 1.3221 & 1.0305 & 0.0064\\ 
            \bottomrule
          \end{tabular}
\end{table}

Providing more contextual information for instructions enhances the performance of MLLMs. 
To further improve MLLM evaluation performance, we enhanced the prompts for both SFT and inference stages. 
As shown in Table ~\ref{tab:instruction_enhancement}, our experiments demonstrate that the enhanced prompts significantly increase evaluation accuracy. Specifically, the evaluation using enhanced instructions reduced the MAE metric by half, from 0.120 to 0.006, compared to the original instructions. 
Additionally, this approach consistently improved evaluation performance across different text-to-image models.

\subsection{Mulit-scaling Resolutions Experiments}

\begin{table}[t]
    \centering
    \small
    \tabcolsep 1.5pt 
    \caption{\small \textbf{Ablation study on multi-scale input.} 
            Results are reported on image faithfulness under different input strategy. We observe that input with multi-scale resolution images can improves the evaluation metrics. MAE: mean absolute error.}
    \label{tab:Resolutions}
    \begin{tabular}{l|c|cccccccc|c}
            \toprule
             Method & Multi Scale   & SDXL & Pixart & Wuerstchen & SDXL-Turbo & IF & SD v1.5 & SD v2.1 & LCM & MAE\\ 
             \midrule
             Human & --                          & 2.1044 & 1.8606 & 1.7839 & 1.3854 & 1.3822 & 1.3818 & 1.1766 & 1.0066 & 0 \\
             \midrule
             \multirow{3}{*}{\ourmethod}  &   & 1.8105 & 1.9238 & 1.9325 & 1.2078 & 1.2247 & 1.4540 & 1.3012 & 1.0554 & 0.1358 \\ 
              &        \checkmark             & 2.0443 & 1.9199 & 1.8012 & 1.3353 & 1.296 & 1.4702 & 1.3221 & 1.0305 & 0.0064 \\ 
            \bottomrule
          \end{tabular}
\end{table}

In the design of LLaVA-Next, using multi-scale resolution images as input helps address the issue of detail information loss, which significantly impacts the evaluation of image faithfulness, such as assessing deformations in hands and faces. We conducted a multi-scale image training comparison experiment to validate this approach. The baseline was the 13B LLaVA model with 336$\times$336 resolution input, while the comparison model used images at three resolutions (336$\times$336, 672$\times$672, 1008$\times$1008) as input. As shown in Table~\ref{tab:Resolutions}, training with multi-scale inputs significantly enhanced the model's understanding of image and achieved better evaluation performance.




\section{Qualitative Analysis}
\begin{figure}[ht]
\centering
\includegraphics[width=\linewidth,keepaspectratio]{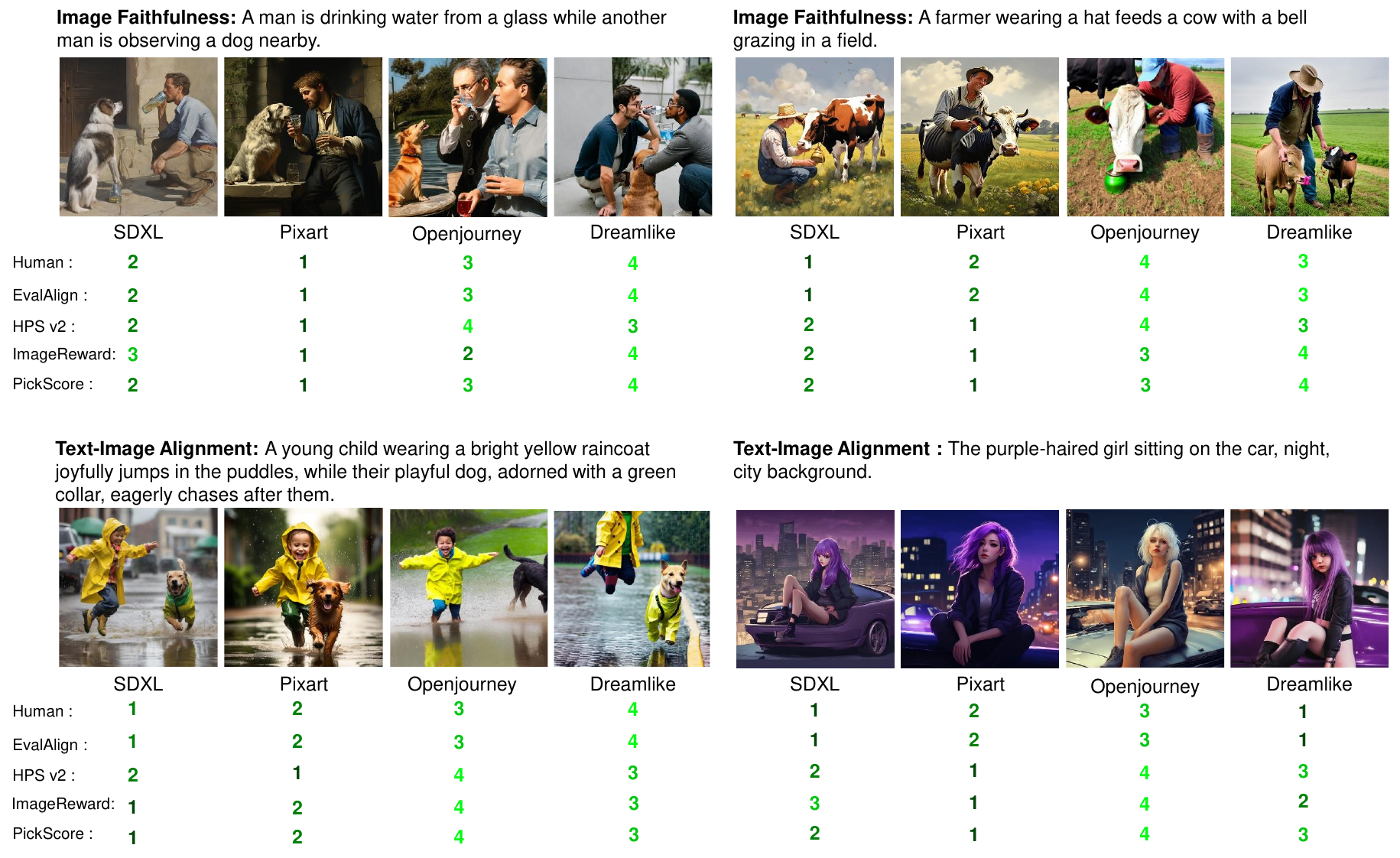}
\vspace{-5pt}
\caption{
    \small \textbf{Qualitative results of \ourmethod dataset and benchmark.} As can be concluded, \ourmethod is consistently aligned with fine-grained human preference in terms of image faithfulness and text-image alignment, while other methods fail to do so.
}
\vspace{-5mm}
\label{fig:qa_comp}
\end{figure}

As shown in Figure~\ref{fig:qa_comp}, we present a comparison of different evaluation metrics on images generated by four models, including human annotated scores, \ourmethod, ImageReward~\citep{xu2024imagereward}, HPSv2~\citep{wu2023hpsv2}, and PickScore~\citep{kirstain2024pickscore}. The digits in the figure represent the ranking for each evaluation metric, with darker colors indicating higher rankings.
From the figure, it is evident that our proposed EvalAlign metric closely matches the human rankings across two evaluation dimensions, demonstrating excellent consistency.
Specifically, the numbers in the figure represent \ourmethod scores for the corresponding evaluation aspect, with darker colors indicating higher scores and better generation performance. Note that if the text prompt does not specify a particular style, the style consistency score defaults to 0.
From these results, it is evident that the same text-to-image model exhibits significant performance variation across different evaluation aspects.

\end{document}